\documentclass[runningheads]{llncs}

\usepackage{graphicx}
\usepackage{amsmath,amssymb,amsfonts}
\usepackage[noend]{algorithmic}
\usepackage{appendix}
\usepackage{makecell}
\usepackage{multirow}
\usepackage{listings}
\usepackage{float}
\usepackage[dvipsnames]{xcolor}
\usepackage{subcaption}
\usepackage[ruled,vlined]{algorithm2e}
\usepackage{tikz}
\usetikzlibrary{decorations.markings}
\usetikzlibrary{positioning}
\usetikzlibrary{calc}
\usepackage{cite}
\usepackage{seqsplit}
\usepackage{enumitem}
\usepackage{hyperref}
\usepackage{changepage}
\hypersetup{ 
  colorlinks,                    
  linkcolor={black},      
  citecolor={blue},              
  urlcolor={blue},      
}

\def\equationautorefname~#1\null{{\color{black}Equation} (#1)\null}

\makeatletter
\g@addto@macro\normalsize{%
  \setlength\abovedisplayskip{-0.5em}
  \setlength\belowdisplayskip{0.5em}
  \setlength\abovedisplayshortskip{-0.5em}
  \setlength\belowdisplayshortskip{0.5em}
}
\makeatother

\DeclareSymbolFont{largesymbolsA}{U}{txexa}{m}{n}
\DeclareMathSymbol{\varprod}{\mathop}{largesymbolsA}{16}

\newcommand{\conlow}[1]{l_{#1}}
\newcommand{\conup}[1]{u_{#1}}

\newcommand{\segmpara}{\sigma}
\newcommand{\symlow}[1]{l^s_{#1}}
\newcommand{\symup}[1]{u^s_{#1}}

\newcommand{\Call}[2]{\textsf{#1}{(#2)}}

\begin{document}

\title{Scalable and Modular Robustness Analysis of Deep Neural Networks}

\author{Yuyi Zhong\inst{1} \quad
  Quang-Trung Ta\inst{1}  \quad
  Tianzuo Luo \inst{1} \\
  Fanlong Zhang \inst{2} \quad
  Siau-Cheng Khoo\inst{1}
  \\\vspace{1em}
  \textup{\footnotesize
    \{yuyizhong, taqt, tianzuoluo\}@comp.nus.edu.sg\\
    izhangfanlong@gmail.com \quad
    khoosc@nus.edu.sg
  }
}

\authorrunning{Yuyi Zhong et al.}
\institute{School of Computing, National University of Singapore, Singapore
  \and School of Computer, Guangdong University of Technology, China
}

\maketitle              

\begin{abstract}
  As neural networks are trained to be deeper and larger, the scalability of
  neural network analyzer is urgently required.
  The main technical insight of our method is modularly analyzing neural
  networks by segmenting a network into blocks and conduct the analysis for each
  block.
  In particular, we propose the {\em network block summarization} technique to
  capture the behaviors within a network block using a block summary and
  leverage the summary to speed up the analysis process.
  We instantiate our method in the context of a CPU-version of the state-of-the-art analyzer DeepPoly and name our system as \emph{Bounded-Block Poly} ({\em
    BBPoly}). We evaluate BBPoly extensively on various experiment settings.
  The experimental result indicates that our method yields comparable precision
  as DeepPoly but runs faster and requires less computational resources.
  Especially, BBPoly can analyze \emph{really} large neural networks like
  SkipNet or ResNet that contain up to one million neurons in less than around 1
  hour per input image, while DeepPoly needs to spend even 40 hours to analyze
  one image. \vspace{-1em}

\end{abstract}


\section{Introduction}

Deep neural networks are one of the most well-established
techniques and have been applied in a wide range of research and engineering
domains such as image classification, autonomous driving etc.
However, researchers have found out that neural nets can sometimes be brittle and show unsafe behaviors.
For instance, a well-trained network may have high accuracy in classifying the
testing image dataset.
But, if the testing image is perturbed subtly without changing the context
of the image, it could fool the network into classifying the perturbed image as
something else; this perturbation is known as adversarial attack
\cite{REN2020346,YuanHZL19}.
To tackle the issue, robustness verification is used to guarantee that unsafe
states will not be reached within a certain perturbation size.
Several verification techniques have been proposed to verify the
robustness of neural networks.

In general, these techniques can be categorized into incomplete methods (e.g.
abstract interpretation \cite{PulinaT10, GehrMDTCV18, SinghGPV19popl}) and
complete methods (e.g. constraint solving \cite{TjengXT19, KatzBDJK17}).
Complete methods reason over exact result, but also require long execution time
and heavy computational power.
On the contrary, incomplete methods run much faster but will lose precision
along the way.

One of the most state-of-the-art neural network verification methods
proposed in recent years is DeepPoly \cite{SinghGPV19popl}.
It is an incomplete but efficient method that uses abstract interpretation
technique to over-approximate operations in neural network.
In particular, DeepPoly designs the abstract domain to
contain symbolic lower and upper constraints, together with concrete lower and
upper bounds of a neuron's value.
The symbolic constraints of a neuron are defined over neurons in the previous layer; during
analysis, they will be revised repeatedly into constraints defined over
neurons of even earlier layers.
This computation is named as \emph{back-substitution} and is aimed to obtain
more precise analysis results \cite{SinghGPV19popl}.

Considering a network with $n$ affine layers and each layer has at most $N$
neurons, the time complexity of this back-substitution operation is $O(n^2\cdot
N^3)$ \cite{MullerSPT20}.
When the neural network has many layers ($n$ is large), this computation is
heavy and it also demands extensive memory space.
This is the main bottleneck of the abstract-interpretation-based analysis used
by DeepPoly.

\textbf{Motivation.} As deep neural networks are trained to be larger and deeper
to achieve higher accuracy or handle more complicated tasks, the verification
tools will inevitably need to scale up so as to analyze more advanced neural
networks.

To mitigate the requirement for high computational power of DeepPoly, we propose
a {\em network block summarization} technique to enhance the scalability of the
verification tool.
Our key insight is to define a method that enables trade-off between precision
requirement, time-efficiency requirement and computing-resource limitation.
Our method, specially tailored to handle very deep networks, leads to faster
analysis and requires less computational resources with reasonable sacrifice of
analysis precision.
We instantiate our method in the context of a CPU-version of DeepPoly, but it
can also be implemented for the GPU version of DeepPoly (named as GPUPoly
\cite{MullerSPT20}) which can lead to even more gain in speed.

\vspace{0.3em}
\textbf{Contribution.} We summarize our contributions below:

\begin{itemize}[noitemsep, topsep=0pt, left=0.5em]
\item We propose {\em block summarization technique} supplemented with bounded back-substitution heuristic to scale up the verification process to handle large networks like ResNet34 \cite{HeZRS16} with around one million neurons.

\item We design two types of block summaries that allow us to take ``shortcuts"
  in the back-substitution process for the purpose of reducing the time
  complexity and memory requirement during the analysis process.

\item We implement our proposal into a prototype analyzer called BBPoly, which
  is built on top of the CPU-version of DeepPoly, and conduct extensive
  experiments on fully-connected, convolutional and residual networks.
  The experimental results show that BBPoly is faster and requires less memory
  allocation compared to the original DeepPoly, while achieves comparable
  precision.
\end{itemize}


\section{Overview}
\label{sec:motivation}
\vspace{-0.5em}
We present an overview of the whole analysis process
with an illustrative example.
Our analyzer is built on top of DeepPoly system, leveraging their design
of abstract domains and abstract transformers.
But we will analyze the network {\em in blocks} and generate block summarization to speed up the analysis process.
Formal details of our proposed method will be provided in
\autoref{sec:summaryFormalDesci}.

\begin{figure*}[htbp]
  \centering
  \begin{tikzpicture}[
      red_node/.style={circle, draw=red, fill=red!5, thin, minimum size = 6mm},
      blue_node/.style={circle, draw = blue, fill=cyan!5, thin, minimum size = 6mm},
      black_node/.style={circle, draw = black, fill=black!5, thin, minimum size = 6mm} ]
    \footnotesize
    \node[red_node] (input1){};
    \node[red, left = 18mm of input1] (interval1) {};
    \node[red, above = 4mm of input1] () {Input layer};
    \node[black, above left = -2mm and 2mm of input1] () {$i_1\in[-1,1]$};
    \node[black, above right = -2.5mm and 1.5mm of input1] () {1};
    \node[black, below right = 2.5mm and 1.5mm of input1] () {1};
    \node[black, below right = 10mm and 1.5mm of input1] () {1};
    \node[black, below right = 18.5mm and 1.5mm of input1] () {-1};
    \node[red_node, below = 14mm of input1](input2){};
    \node[black, above left = -2mm and 2mm of input2] () {$i_2\in[-1,1]$};
    \node[red, left = 18mm of input2] (interval2) {};
    \node[black_node, right = 18mm of input1](x1){};
    \node[black, above right = -2.5mm and 1.5mm of x1] () {1};
    \node[black, below right = 2.5mm and 1.5mm of x1] () {1};
    \node[black, below right = 10mm and 1.5mm of x1] () {1};
    \node[black, below right = 18.5mm and 1.5mm of x1] () {-1};
    \node[black, above = 0mm of x1] () {0};
    \node[black, above = 4mm of x1] () {Hidden layer 1};
    \node[black_node, below = 14mm of x1](x2){};
    \node[black, below = 0mm of x2] () {0};
    \node[black_node, right = 18mm of x1](x3){};
    \node[black, above right = -2.5mm and 1.5mm of x3] () {1};
    \node[black, below right = 2.5mm and 1.5mm of x3] () {0};
    \node[black, below right = 10mm and 1.5mm of x3] () {1};
    \node[black, below right = 18.5mm and 1.5mm of x3] () {1};
    \node[black, above = 0mm of x3] () {0};
    \node[black, above = 4mm of x3] () {Hidden layer 2};
    \node[black_node, below = 14mm of x3](x4){};
    \node[black, below = 0mm of x4] () {0};
    \node[blue_node, right = 18mm of x3](output1){};
    \node[black, above = 0mm of output1] () {1};
    \node[blue, above = 4mm of output1] () {Output layer};
    \node[blue_node, below = 14mm of output1](output2){};
    \node[black, below = 0mm of output2] () {0};
    \draw [->,black,thin](interval1) -- (input1);
    \draw [->,black,thin](interval2) -- (input2);
    \draw [->,black,thin](input1) -- (x1);
    \draw [->,black,thin](input2) -- (x2);
    \draw [->,black,thin](input1) -- (x2);
    \draw [->,black,thin](input2) -- (x1);
    \draw [->,black,thin](x1) -- (x3);
    \draw [->,black,thin](x2) -- (x4);
    \draw [->,black,thin](x1) -- (x4);
    \draw [->,black,thin](x2) -- (x3);
    \draw [->,black,thin](x3) -- (output1);
    \draw [->,black,thin](x4) -- (output2);
    \draw [->,black,thin](x3) -- (output2);
    \draw [->,black,thin](x4) -- (output1);
  \end{tikzpicture}

  \caption{Example fully-connected network with ReLU activation (cf. \cite{SinghGPV19popl})}
  \label{fig:running_example}

\end{figure*}
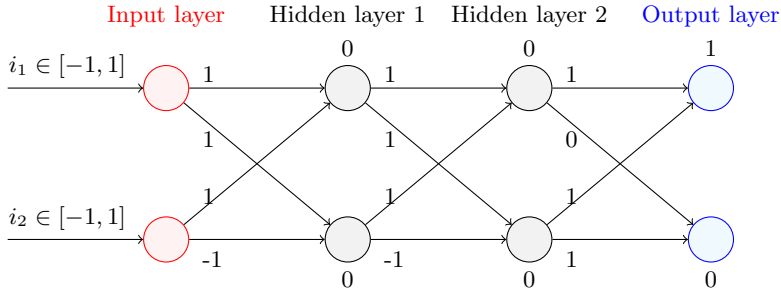

The illustrative example is a fully-connected network with ReLU activation
function as shown in \autoref{fig:running_example}.
The network has 4 layers with 2 neurons in each layer and the two input neurons
$i_1, i_2$ can independently take any real number between $[-1, 1]$.
The weights of the connections between any two neurons from two adjacent layers
are displayed at their corresponding edges, the bias of each neuron is indicated
either above or below the neuron.
Computation for a neuron in a hidden layer undergoes two steps: (i) an
{\em affine transformation\/} based on the inputs, weights and biases related to
this neuron, which generates a value $v$, followed by (ii) a {\em ReLU
  activation} which outputs $v$ if $v>0$, or 0 if $v \leq 0$.
For the output layer, only affine transformation is applied to generate the
final output of the entire network.

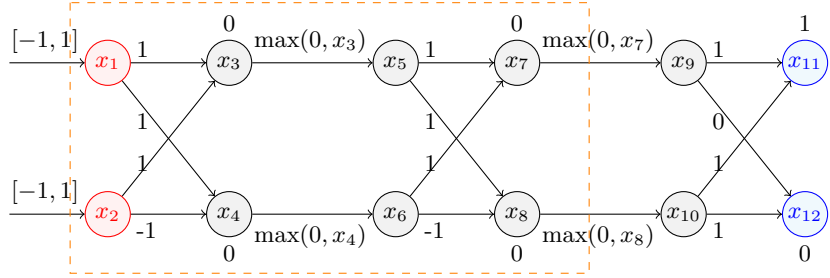
\begin{figure}[htbp]
\centering
\begin{tikzpicture}[
    red_node/.style={circle, draw=red, fill=red!5, thin,
      minimum size = 6mm, inner sep=1pt},
    blue_node/.style={circle, draw = blue, fill=cyan!5, thin,
      minimum size = 6mm, inner sep=1pt},
    black_node/.style={circle, draw = black, fill=black!5, thin,
      minimum size = 6mm, inner sep=1pt},
    red_rectangle/.style={rectangle, draw = red, thin, dashed}
    ]
  \node[red, red_node] (x1){$x_1$};
  \node[red, left = 10mm of x1] (interval1) {};
  \node[black, above left = -2mm and 0.5mm of x1] () {$[-1,1]$};
  \node[black, above right = -2.5mm and 0.5mm of x1] () {1};
  \node[black, below right = 3.5mm and 0.5mm of x1] () {1};
  \node[black, below right = 9mm and 0.5mm of x1] () {1};
  \node[black, below right = 18mm and 0.5mm of x1] () {-1};
  \node[red, red_node, below = 14mm of x1](x2){$x_2$};
  \node[black, above left = -2mm and 0.5mm of x2] () {$[-1,1]$};
  \node[red, left = 10mm of x2] (interval2) {};
  \node[black_node, right = 10mm of x1](x3){$x_3$};
  \node[black, above = 0mm of x3] () {0};
  \node[black, above right = -2 mm and 0 mm of x3] () {$\mathrm{max}(0, x_3)$};
  \node[black_node, below = 14mm of x3](x4){$x_4$};
  \node[black, below = 0mm of x4] () {0};
  \node[black, below right = -2 mm and 0 mm of x4] () {$\mathrm{max}(0, x_4)$};
  \node[black_node, right = 16mm of x3](x5){$x_5$};
  \node[black_node, below = 14mm of x5](x6){$x_6$};
  \node[black_node, right = 10mm of x5](x7){$x_7$};
  \node[black_node, below = 14mm of x7](x8){$x_8$};
  \node[black, above right = -2.5mm and 0.5mm of x5] () {1};
  \node[black, below right = 3.5mm and 0.5mm of x5] () {1};
  \node[black, below right = 9mm and 0.5mm of x5] () {1};
  \node[black, below right = 18mm and 0.5mm of x5] () {-1};
  \node[black, above = 0mm of x7] () {0};
  \node[black, below = 0mm of x8] () {0};
  \node[black_node, right = 16mm of x7](x9){$x_9$};
  \node[black_node, below = 14mm of x9](x10){$x_{10}$};
  \node[black, above right = -2 mm and 0 mm of x7] () {$\mathrm{max}(0, x_7)$};
  \node[black, below right = -2 mm and 0 mm of x8] () {$\mathrm{max}(0, x_8)$};
  \node[blue, blue_node, right = 10mm of x9](x11){$x_{11}$};
  \node[blue, blue_node, below = 14mm of x11](x12){$x_{12}$};
  \node[black, above right = -2.5mm and 0.5mm of x9] () {1};
  \node[black, below right = 3.5mm and 0.5mm of x9] () {0};
  \node[black, below right = 9mm and 0.5mm of x9] () {1};
  \node[black, below right = 18mm and 0.5mm of x9] () {1};
  \node[black, above = 0mm of x11] () {1};
  \node[black, below = 0mm of x12] () {0};
  \draw [->,black,thin](interval1) -- (x1);
  \draw [->,black,thin](interval2) -- (x2);
  \draw [->,black,thin](x1) -- (x3);
  \draw [->,black,thin](x2) -- (x3);
  \draw [->,black,thin](x1) -- (x4);
  \draw [->,black,thin](x2) -- (x4);
  \draw [->,black,thin](x3) -- (x5);
  \draw [->,black,thin](x4) -- (x6);
  \draw [->,black,thin](x5) -- (x7);
  \draw [->,black,thin](x6) -- (x8);
  \draw [->,black,thin](x5) -- (x8);
  \draw [->,black,thin](x6) -- (x7);
  \draw [->,black,thin](x7) -- (x9);
  \draw [->,black,thin](x8) -- (x10);
  \draw [->,black,thin](x9) -- (x11);
  \draw [->,black,thin](x10) -- (x11);
  \draw [->,black,thin](x9) -- (x12);
  \draw [->,black,thin](x10) -- (x12);
  \draw[orange, thin, dashed] (-0.5,0.8) rectangle (6.4,-2.8);
\end{tikzpicture}
\vspace{0.3em}
\caption{The transformed network from \autoref{fig:running_example}
  to perform analysis (cf. \cite{SinghGPV19popl})}
\label{fig:HypoNet}
\end{figure}

\vspace{-0.5em}
To analyze a neural network, we follow the approach taken by DeepPoly where each hidden layer is perceived as a combination of an affine layer and a ReLU layer.
Therefore, network in \autoref{fig:running_example} will be represented by the
network depicted in \autoref{fig:HypoNet} for analysis purpose, where a neuron
in a hidden layer is expanded into two nodes: (i) one affine node for the
related affine transformation (such as $x_3,x_4, x_7, x_8)$, and (ii) one ReLU
node which is the output of ReLU function (such as $x_5. x_6, x_9,x_{10}$).

\vspace{-2em}
\subsection{Preliminary Description on Abstract Domain}
\label{des_on_abs_domain}

We use the abstract
domain designed from DeepPoly system \cite{SinghGPV19popl} to verify neural networks.
For each neuron $x_i$, its abstract value is comprised of four elements: a
symbolic upper constraint $\symup{i}$, a symbolic lower constraint
$\symlow{i}$, a concrete lower bound $l_i$ and a concrete upper bound $u_i$.
And we have $\symlow{i} \leq x_i \leq \symup{i}$, $x_i \in [l_i,u_i]$.
All the symbolic constraints associated with $x_i$ can be formulated as
$b_i+\sum_j{w_j\cdot x_j}$, where $w_j \in \mathbb{R}, b_i \in \mathbb{R}, j<i$.
Here, the constraint $j<i$ asserts that the constraints for $x_i$ only refer to
variables ``before'' $x_i$, since the value of one neuron (at a layer) only
depends on the values of the neurons at preceding layers.
For the concrete bounds of $x_i$, we have $l_i \in \mathbb{R}, u_i \in
\mathbb{R}, l_i \leq u_i$ and the interval $[l_i,u_i]$
over-approximates all the values that $x_i$ can possibly take.
\vspace{-1em}
\subsection{Abstract Interpretation on the Example Network}
\label{sec:abInt_on_egNet}

\vspace{-0.5em}
We now illustrate how to apply abstract interpretation on the example network
in order to get the output range of the network, given an abstract input $[-1,1]$
for both the input neurons.

The analysis starts at the input layer and processes layer by layer until output layer. The abstract values of the inputs $x_1,x_2$ are
respectively $\langle \symlow{1} =-1, \symup{1} =1, \conlow{1} = -1, \conup{1} =
1\rangle$ and $\langle \symlow{2} =-1, \symup{2} =1, \conlow{2} = -1, \conup{2}
= 1 \rangle$. Next, the affine abstract transformer (designed by DeepPoly \cite{SinghGPV19popl}) for $x_3$ and $x_4$
generates the following symbolic constraints, where the coefficients (and the
constant terms, if any) in constraints are the weights (and bias) in the fully
connected layer:

\begin{equation}
\label{equ:x12_syb_cons}
\begin{split}
  x_1 + x_2 \leq x_3 \leq x_1+x_2;\quad
  \ x_1-x_2 \leq x_4 \leq x_1-x_2 \\
\end{split}
\end{equation}

The concrete bounds are computed using concrete intervals of $x_1, x_2$ and symbolic constraints in \autoref{equ:x12_syb_cons}, thus $l_3=l_4=-2$ and $u_3=u_4=2$ (the process of computing concrete bound is formally described in Appendix \ref{appendix:concre_bound_computation}).

The activation transformer (designed by DeepPoly \cite{SinghGPV19popl}) is then applied to get the abstract elements for
$x_5,x_6$ from $x_3, x_4$ respectively.
In general, given that $x_i = \mathrm{ReLU}(x_j)$, if $u_j\leq 0$, $x_i$ is
always 0, therefore we have $0\leq x_i \leq 0, l_i=0,
u_i =0$. If $l_j\geq 0$, then $x_i = x_j$ and we get $x_j\leq x_i\leq x_j,
l_i=l_j, u_i =u_j$. For the case where $l_{j}<0$ and  $u_{j}>0$, an
over-approximation error will be introduced and we set the abstract element as
followed for $x_i$:

\begin{equation}\label{equ:relu_tri_form}
\begin{gathered}
  x_i \geq c_i\cdot x_j,
  \quad \ x_i \leq \frac{u_j(x_j-l_j)}{u_j-l_j},
  \quad \ l_i = 0,
  \quad \ u_i = u_j,
\end{gathered}
\end{equation}

where $c_i =0$ if $|l_j| > |u_j|$ and $c_i=1$  otherwise. For example, $x_5 =
\mathrm{ReLU}(x_3)$ and since $l_3 <0, u_3>0$, it belongs to the last case
described in \autoref{equ:relu_tri_form}. $|l_3| = |u_3| =2$ therefore $c_5=1$.
Finally, we get the abstract value for $x_5$: $l_5=0, u_5=2, \symlow{5}=x_3,
\symup{5} = 0.5\cdot x_3 +1$. Similar computation can be done for $x_6$ to yield
$l_6=0, u_6=2, \symlow{6}=x_4, \symup{6} = 0.5\cdot x_4 +1$.

Next, we work on the symbolic bounds for $x_7, x_8$, beginning with:

\begin{equation}
\label{equ:x78_syb_cons}
\begin{split}
  x_5 + x_6 \leq x_7 \leq x_5+x_6;
  \quad \ x_5-x_6 \leq x_8 \leq x_5-x_6
\end{split}
\end{equation}

From the symbolic constraints in \autoref{equ:x78_syb_cons}, we recursively substitute the symbolic constraints {\em backward} layer by layer until the constraints are expressed in terms of the input variables. Upon reaching back to an earlier layer, constraints defined over neurons in that layer are constructed and concrete bound values are evaluated and recorded (refer to Appendix \ref{appendix:concre_bound_computation}). Finally the most precise bound among all these layers will be selected as the actual concrete bound for $x_7$ and $x_8$ respectively. This process is called {\em back-substitution} and is the key technique proposed in DeepPoly to achieve tighter bounds. We follow the back-substitution procedure in DeepPoly and construct constraints for $x_7,x_8$ defined over $x_3, x_4$:

\begin{equation}
\label{equ:x78_define_over_x34}
\begin{gathered}
  x_3+x_4 \leq x_7 \leq 0.5\cdot x_3+0.5\cdot x_4+2
  \\
  x_3-(0.5\cdot x_4+1)\leq x_8 \leq 0.5\cdot x_3+1-x_4 ,
\end{gathered}
\end{equation}

And we further back-substitute to have them defined over $x_1$, $x_2$:

\begin{equation}
\label{equ:x78_define_over_x12}
\begin{gathered}
     2x_1 \leq x_7 \leq x_1+2\\
      0.5\cdot x_1 +1.5 \cdot x_2-1 \leq x_8 \leq -0.5\cdot x_1+1.5 \cdot x_2 +1
\end{gathered}
\end{equation}

Finally, we determine the best bound for $x_7$ to be $l_7=0, u_7=3$ and that for $x_8$ to be $l_8=-2, u_8=2$.
Note that we have additionally drawn a dashed orange box in
\autoref{fig:HypoNet} to represent a network {\em block}.
Here, we propose a {\em block summarization} method which captures the relations
between the input (leftmost) layer and output (rightmost) layer of the block.
Thus \autoref{equ:x78_define_over_x12} can function as the block summarization for
the dashed block in \autoref{fig:HypoNet}; we  leverage on this block summarization to make ``jumps" during back-substitution process so as to save both
running time and memory (details in \autoref{sec:summaryFormalDesci}).

To continue with our analysis process, we obtain next:

\begin{equation}
\label{equ:x8910}
\begin{gathered}
   l_9=0,\quad u_9=3,\quad \symlow{9}=x_7,\quad \symup{9}=x_7\\
    l_{10}=0,\quad u_{10}=2,\quad \symlow{10}=x_8,\quad \symup{10} = 0.5\cdot x_8 +1\\
    l_{11}=1,\quad u_{11}=6,\quad \symlow{11}=x_9+x_{10}+1,\quad \symup{11}= x_9+x_{10}+1\\
    l_{12}=0,\quad u_{12}=2,\quad \symlow{12}=x_{10},\quad \symup{12}= x_{10},
\end{gathered}
\end{equation}

Here, we can quickly construct the constraints of $x_{11}$ defined over
$x_1,x_2$ by using the block summarization derived in \autoref{equ:x78_define_over_x12}; yielding $2.5\cdot x_1+1.5\cdot x_2 \leq
x_{11}\leq 0.75\cdot x_1 +0.75\cdot x_2 +4.5$. By doing so,
our analysis will return $x_{11} \in [1, 6]$ and $x_{12} \in
[0,2]$.
Note that we lose some precision when making ``jumps" through block
summarization; the interval for $x_{11}$ would be $[1, 5.5]$ if we were to stick to
layer-by-layer back-substitution as originally designed in DeepPoly.

\vspace{-1em}
\subsection{Scaling up with block summarization}

As illustrated in \autoref{equ:x78_define_over_x34} and
\autoref{equ:x78_define_over_x12}, we conduct back-substitution to construct
symbolic constraints defined over neurons at earlier layer in order to obtain a tighter concrete bound.
In DeepPoly, every affine layer
initiates layer-by-layer back-substitution until the input layer.
Specifically, we assume a network with $n$ affine layers, maximum $N$ neurons per layer and consider the $k^{\mathit{th}}$ affine layer (where the input layer is indexed as $0$).
Every step of back-substitution for layer $k$ through a preceding
affine layer requires $O(N^3)$ time complexity and every back-substitution through a preceding ReLU layer requires $O(N^2)$, it takes $O(k \cdot N^3)$ for the $k^{\mathit{th}}$ affine layer to complete the back-substitution process.
Overall, DeepPoly analysis requires $O(n^2\cdot N^3)$ time complexity.
This can take a toll on DeepPoly when handling large networks.
For example, in our evaluation platform, DeepPoly takes around 40 hours to
analyze one image on ResNet18 \cite{HeZRS16} with 18 layers.
Therefore, we propose to divide the neural networks into blocks, and compute
the summarization for each block.
This summarization enables us to charge up the back-substitution operation by speeding across blocks, as demonstrated in \autoref{equ:x78_define_over_x12} where constraints of neuron $x_{7}$ are directly defined over input neurons.
%

\vspace{-1em}

\section{Network Block Summarization}
\label{sec:summaryFormalDesci}
\vspace{-0.5em}
\subsection{Network analysis with modularization}\label{sec:form_modul}

For better scalability, we propose a modularization methodology to decrease the
computational cost as well as the memory usage, where we segment the network
into blocks and analyze each block in sequence.
Specifically, we propose the following two techniques to reduce computation
steps:

\begin{enumerate}[topsep=0pt]
\item Generate summarization between the input and output neurons for each
  block, and leverage block summarization to make ``jumps'' during
  back-substitution instead of doing it layer by layer.

\item Leverage block summarization by bounding back-substitution operation to
  terminate early.
\end{enumerate}
\vspace{-0.5em}

\begin{figure*}[t!]
  \centering
  \begin{subfigure}[b]{0.5\textwidth}
    \centering
    \includegraphics[width = \columnwidth]{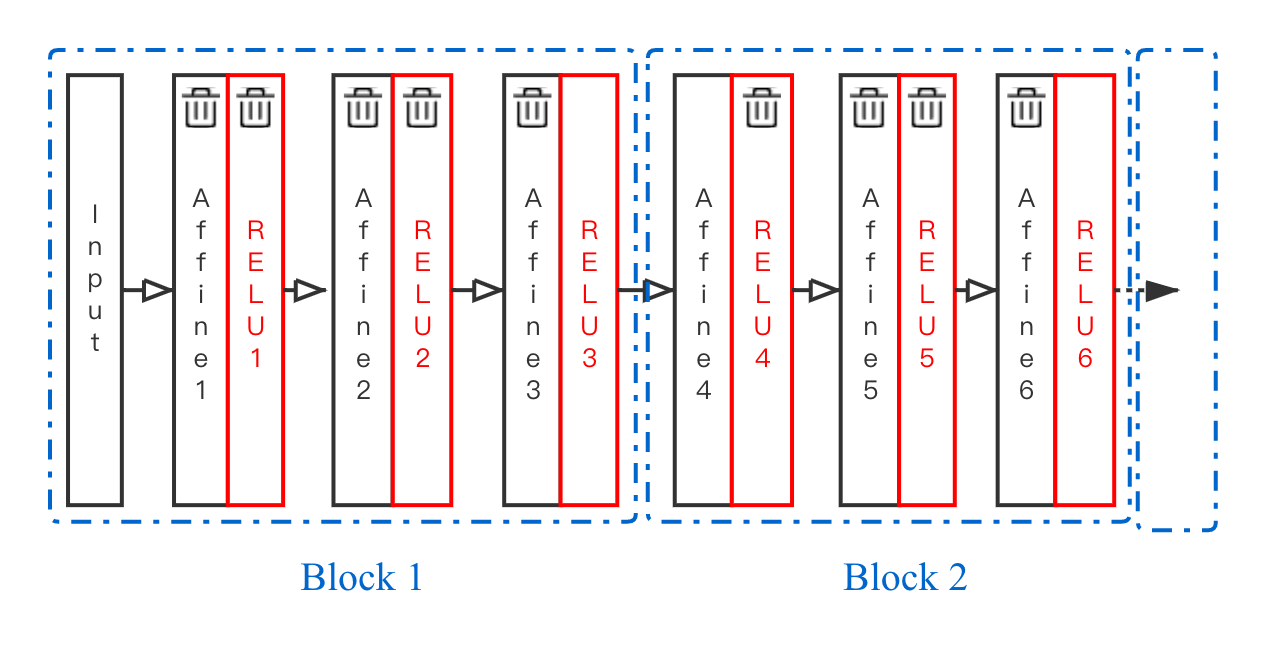}
    \vspace{-2em}
    \caption{Intuitive segmentation}
    \label{fig:intui_segm}
  \end{subfigure}%
  \begin{subfigure}[b]{0.5\textwidth}
    \centering
    \includegraphics[width = \columnwidth]{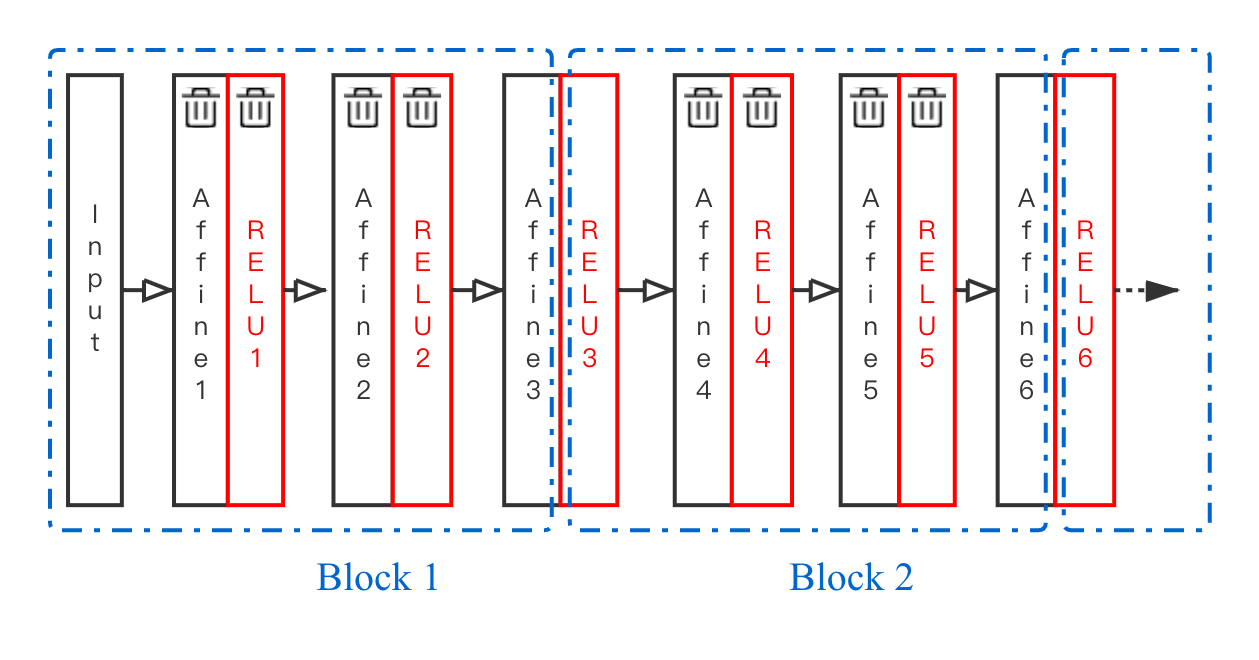}
    \vspace{-2em}
    \caption{Designed segmentation}
    \label{fig:final_seg}
  \end{subfigure}
  \caption{Example on Segmenting Network Into Blocks}
  \label{fig:net_in_blks}
\end{figure*}

As illustrated by the simplified network representation in
\autoref{fig:net_in_blks}, we segment a network into two blocks.
We then show (1) how to generate summarization given the network fragment and
(2) how to leverage the summarization to perform back-substitution.
The details are as follows.

\begin{figure}[t!]
  \centering
  \def\resNetLayerName{\texttt{\seqsplit{3*3}\\[0.5em]
      \seqsplit{conv}\\[0.5em]
      \seqsplit{64}}}
  \def\resNetonetwoLayerName{\texttt{\seqsplit{3*3}\\[0.5em]
      \seqsplit{conv}\\[0.5em]
      \seqsplit{128}\\[0.5em]
      \seqsplit{/2}}}
  \def\resNetoneLayerName{\texttt{\seqsplit{3*3}\\[0.5em]
      \seqsplit{conv}\\[0.5em]
      \seqsplit{128}}}
  \def\resNettwotwoLayerName{\texttt{\seqsplit{3*3}\\[0.5em]
      \seqsplit{conv}\\[0.5em]
      \seqsplit{256}\\[0.5em]
      \seqsplit{/2}}}
  \def\resNettwoLayerName{\texttt{\seqsplit{3*3}\\[0.5em]
      \seqsplit{conv}\\[0.5em]
      \seqsplit{256}}}
  \def\resNetthreetwoLayerName{\texttt{\seqsplit{3*3}\\[0.5em]
      \seqsplit{conv}\\[0.5em]
      \seqsplit{512}\\[0.5em]
      \seqsplit{/2}}}
  \def\resNetthreeLayerName{\texttt{\seqsplit{3*3}\\[0.5em]
      \seqsplit{conv}\\[0.5em]
      \seqsplit{512}}}
  \def\InputlayerName{\texttt{
      \seqsplit{Input}\\[0.8em]
      \seqsplit{layer}}}
  \def\convLayerName{\texttt{\seqsplit{7*7}\\[0.5em]
      \seqsplit{conv}\\[0.5em]
      \seqsplit{64}\\[0.5em]
      \seqsplit{/2}}}
  \def\poolLayerName{\texttt{\seqsplit{3*3}\\[0.5em]
      \seqsplit{pool}\\[0.5em]
      \seqsplit{/2}}}
  \def\avepoolLayerName{\texttt{\seqsplit{average}\\[0.8em]
      \seqsplit{pool}}}
  \def\fcLayerName{\texttt{\seqsplit{fully}\\[0.8em]
      \seqsplit{connected}}}
  \def\outputLayerName{\texttt{\seqsplit{Output}\\[0.8em]
      \seqsplit{layer}}}
  \renewcommand{\baselinestretch}{0.68}
  \begin{tikzpicture}[
      layer/.style={rectangle, draw=black, thin, text width = 0.5em,
        inner sep=2.3pt, scale=0.8, minimum height=13em},
      decoration={markings, mark= at position 3.6cm with {\arrow{latex}}}
      ]
    \tikzset{>=latex}
    \footnotesize

    \node[layer] (input){\InputlayerName};
    \node[layer, right=0.5em of input] (conv1){\convLayerName};
    \node[layer, right=0.5em of conv1] (pool1){\poolLayerName};
    \node[layer, right=1.2em of pool1] (L1){\resNetLayerName};
    \node[layer, right=0.5em of L1] (L2){\resNetLayerName};
    \node[layer, right=1.4em of L2] (L3){\resNetLayerName};
    \node[layer, right=0.5em of L3] (L4){\resNetLayerName};
    \node[layer, right=1.4em of L4] (L5){\resNetonetwoLayerName};
    \node[layer, right=0.5em of L5] (L6){\resNetoneLayerName};
    \node[layer, right=1.4em of L6] (L7){\resNetoneLayerName};
    \node[layer, right=0.5em of L7] (L8){\resNetoneLayerName};
    \node[layer, right=1.4em of L8] (L9){\resNettwotwoLayerName};
    \node[layer, right=0.5em of L9] (L10){\resNettwoLayerName};
    \node[layer, right=1.4em of L10] (L11){\resNettwoLayerName};
    \node[layer, right=0.5em of L11] (L12){\resNettwoLayerName};
    \node[layer, right=1.4em of L12] (L13){\resNetthreetwoLayerName};
    \node[layer, right=0.5em of L13] (L14){\resNetthreeLayerName};
    \node[layer, right=1.4em of L14] (L15){\resNetthreeLayerName};
    \node[layer, right=0.5em of L15] (L16){\resNetthreeLayerName};
    \node[layer, right=1.2em of L16] (L17){\avepoolLayerName};
    \node[layer, right=0.5em of L17] (L18){\fcLayerName};
    \node[layer, right=0.5em of L18] (L19){\outputLayerName};

    \draw[->] (input.east) -- (conv1.west);
    \draw[->] (conv1.east) -- (pool1.west);
    \draw[->] (pool1.east) -- (L1.west);
    \draw[->] (L1.east) -- (L2.west);
    \draw[->] (L2.east) -- (L3.west);
    \draw[->] (L3.east) -- (L4.west);
    \draw[->] (L4.east) -- (L5.west);
    \draw[->] (L5.east) -- (L6.west);
    \draw[->] (L6.east) -- (L7.west);
    \draw[->] (L7.east) -- (L8.west);
    \draw[->] (L8.east) -- (L9.west);
    \draw[->] (L9.east) -- (L10.west);
    \draw[->] (L10.east) -- (L11.west);
    \draw[->] (L11.east) -- (L12.west);
    \draw[->] (L12.east) -- (L13.west);
    \draw[->] (L13.east) -- (L14.west);
    \draw[->] (L14.east) -- (L15.west);
    \draw[->] (L15.east) -- (L16.west);
    \draw[->] (L16.east) -- (L17.west);
    \draw[->] (L17.east) -- (L18.west);
    \draw[->] (L18.east) -- (L19.west);

    \node (I1A) at ($(pool1)!0.36!(L1)$) {};
    \node (I1B) at ($(pool1)!0.63!(L1)$) {};
    \node (I2A) at ($(L2)!0.36!(L3)$) {};
    \node (I2B) at ($(L2)!0.63!(L3)$) {};
    \node (I3A) at ($(L4)!0.36!(L5)$) {};
    \node (I3B) at ($(L4)!0.63!(L5)$) {};
    \node (I4A) at ($(L6)!0.36!(L7)$) {};
    \node (I4B) at ($(L6)!0.63!(L7)$) {};
    \node (I5A) at ($(L8)!0.36!(L9)$) {};
    \node (I5B) at ($(L8)!0.63!(L9)$) {};
    \node (I6A) at ($(L10)!0.36!(L11)$) {};
    \node (I6B) at ($(L10)!0.63!(L11)$) {};
    \node (I7A) at ($(L12)!0.36!(L13)$) {};
    \node (I7B) at ($(L12)!0.63!(L13)$) {};
    \node (I8A) at ($(L14)!0.36!(L15)$) {};
    \node (I8B) at ($(L14)!0.63!(L15)$) {};
    \node (I9A) at ($(L16)!0.36!(L17)$) {};
    \node (I9B) at ($(L16)!0.63!(L17)$) {};

    \node (C1) at ($(I1B)!0.5!(I2A)$) {};
    \node (C2) at ($(I2B)!0.5!(I3A)$) {};
    \node (C3) at ($(I3B)!0.5!(I4A)$) {};
    \node (C4) at ($(I4B)!0.5!(I5A)$) {};
    \node (C5) at ($(I5B)!0.5!(I6A)$) {};
    \node (C6) at ($(I6B)!0.5!(I7A)$) {};
    \node (C7) at ($(I7B)!0.5!(I8A)$) {};
    \node (C8) at ($(I8B)!0.5!(I9A)$) {};

    \node (C0) at ($(input)!0.5!(pool1)$) {};
     \node[blue, below = 6em of C0] (){\scriptsize block 1};
     \node[blue, below = 6em of C1] (){\scriptsize block 2};
      \node[blue, below = 6em of C2] (){\scriptsize block 3};
    \node[blue, below = 6em of C3] (){\scriptsize block 4};
     \node[blue, below = 6em of C4] (){\scriptsize block 5};
     \node[blue, below = 6em of C5] (){\scriptsize block 6};
     \node[blue, below = 6em of C6] (){\scriptsize block 7};
     \node[blue, below = 6em of C7] (){\scriptsize block 8};
     \node[blue, below = 6em of C8] (){\scriptsize block 9};
     \node (C9) at ($(L17)!0.5!(L19)$) {};
    \node[blue, below = 6em of C9] (){\scriptsize block 10};

    \draw[->] (I1B.center)
    ..controls ($(C1.center) + (-0.6,3.5)$) and ($(C1.center) + (0.6,3.5)$) ..
    (I2A.center);

    \draw[->] (I2B.center)
    ..controls ($(C2.center) + (-0.6,3.5)$) and ($(C2.center) + (0.6,3.5)$) ..
    (I3A.center);

    \draw[dashed, ->] (I3B.center)
    ..controls ($(C3.center) + (-0.6,3.5)$) and ($(C3.center) + (0.6,3.5)$) ..
    (I4A.center);

    \draw[->] (I4B.center)
    ..controls ($(C4.center) + (-0.6,3.5)$) and ($(C4.center) + (0.6,3.5)$) ..
    (I5A.center);

    \draw[dashed, ->] (I5B.center)
    ..controls ($(C5.center) + (-0.6,3.5)$) and ($(C5.center) + (0.6,3.5)$) ..
    (I6A.center);

    \draw[->] (I6B.center)
    ..controls ($(C6.center) + (-0.6,3.5)$) and ($(C6.center) + (0.6,3.5)$) ..
    (I7A.center);

    \draw[dashed, ->] (I7B.center)
    ..controls ($(C7.center) + (-0.6,3.5)$) and ($(C7.center) + (0.6,3.5)$) ..
    (I8A.center);

    \draw[->] (I8B.center)
    ..controls ($(C8.center) + (-0.6,3.5)$) and ($(C8.center) + (0.6,3.5)$) ..
    (I9A.center);


    \draw[blue, thin, dotted]
    ($(C1.center) + (-2.02,2.8)$)
    rectangle
    ($(C1.center) + (-0.65,-2)$);

    \draw[blue, thin, dotted]
    ($(C1.center) + (-0.54,2.8)$)
    rectangle
    ($(C1.center) + (0.54,-2)$);

    \draw[blue, thin, dotted]
    ($(C2.center) + (-0.54,2.8)$)
    rectangle
    ($(C2.center) + (0.54,-2)$);

    \draw[blue, thin, dotted]
    ($(C3.center) + (-0.54,2.8)$)
    rectangle
    ($(C3.center) + (0.54,-2)$);

    \draw[blue, thin, dotted]
    ($(C4.center) + (-0.54,2.8)$)
    rectangle
    ($(C4.center) + (0.54,-2)$);

    \draw[blue, thin, dotted]
    ($(C5.center) + (-0.54,2.8)$)
    rectangle
    ($(C5.center) + (0.54,-2)$);

    \draw[blue, thin, dotted]
    ($(C6.center) + (-0.54,2.8)$)
    rectangle
    ($(C6.center) + (0.54,-2)$);

    \draw[blue, thin, dotted]
    ($(C7.center) + (-0.54,2.8)$)
    rectangle
    ($(C7.center) + (0.54,-2)$);

    \draw[blue, thin, dotted]
    ($(C8.center) + (-0.54,2.8)$)
    rectangle
    ($(C8.center) + (0.54,-2)$);

    \draw[blue, thin, dotted]
    ($(C8.center) + (2.02,2.8)$)
    rectangle
    ($(C8.center) + (0.65,-2)$);
  \end{tikzpicture}
  \vspace{-0.5em}
  \caption{ResNet18 \cite{HeZRS16} and the corresponding block segmentation}
  \vspace{-1.5em}
  \label{fig:resnetseg}
\end{figure}
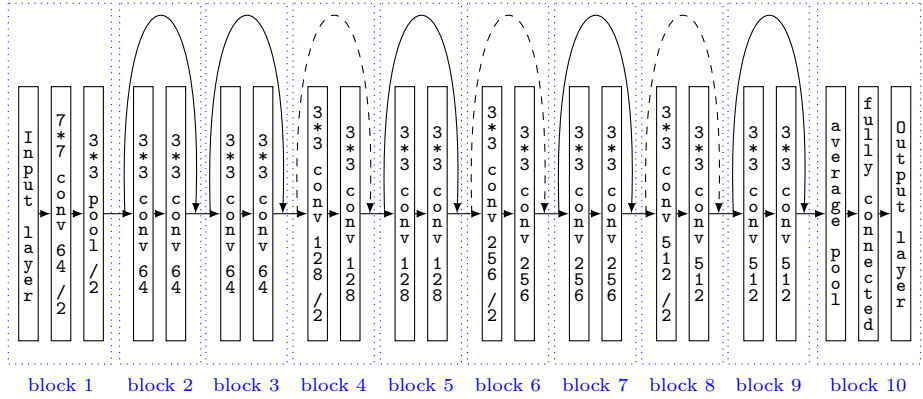

\textbf{Network segmentation.}\label{sec:networksegmentation}
We parameterize network segmentation with a parameter $\segmpara$, which is the number
of affine layers required to form a block.
For example, $\segmpara$ is set to 3 in \autoref{fig:net_in_blks}.
Since each layer in the neural network ends with the ReLU function, an
intuitive segmentation solution is to divide the network so that each block
always {\em ends at a ReLU layer}, as depicted in \autoref{fig:intui_segm}.
However, doing so requires more computation during the back-substitution but
does not gain more accuracy as compared to the segmentation method in which {\em each block ends by an affine layer.}\footnote{An explanation of our choice to end blocks at an affine layer instead of a ReLU layer can be found in Appendix \ref{appendix:explain-layer-end-affine}}
Therefore, we choose the later segmentation option, as shown in
\autoref{fig:final_seg}.

Moreover, special care is required to segment a residual network.
As illustrated in \autoref{fig:resnetseg}, the most important feature of
residual network is the \textit{skip connection} that enables a layer to take
``shortcut connections'' with a much earlier layer \cite{HeZRS16} (displayed by
the curved line).
Thus, a set of layers residing in between a layer and its skip-connected layer
forms an ``intrinsic'' residual block, to be used to segment the network (eg.,
blocks \#2 to \#9 in \autoref{fig:resnetseg}).
A more dynamic choice of block size or block number could potentially lead to
better trade-off between speed and precision; we leave it as future work.

\textbf{Back-substitution with block summarization.}
We present the analysis procedure which implements our block summarization
method (\autoref{sec:summary_in_block}) and bounded back-substitution heuristic
in \autoref{algo:methodalgo}.
Given an input neural network, it will be first segmented (line 1) using the
method described in previous subsection.
For a block consisting of layers $\gamma_a,\ldots,\gamma_k$, the start layer and
the end layer of the block will be remembered by the markers
\textsf{GetStartLayer} and \textsf{IsEndLayer} respectively.
The analysis of ReLU layers (line 3) only depends on the preceding affine layer
it connects to (line 4). The computation of ReLU layer (line 5) follows the
process described in \autoref{sec:abInt_on_egNet} and
\autoref{equ:relu_tri_form}.

To handle affine layer with back-substitution, we firstly assign $\gamma_{pre}$
to be the preceding layer of $\gamma_k$ (line 7).
Then, we initialize the constraint set of $\gamma_k$ to be the symbolic lower
and upper constraints for neurons in $\gamma_k$ (line 8).
Constraints $\Upsilon_k$ are defined over neurons in layer $\gamma_{pre}$ and
directly record the affine transformation between layer $\gamma_k$ and
$\gamma_{pre}$.
Thus, we could use $\Upsilon_k$ and the concrete bounds of neurons in
$\gamma_{pre}$ to compute the initial concrete bounds for neurons in layer
$\gamma_k$ (line 9), using the constraint evaluation mechanism described in
Appendix \ref{appendix:concre_bound_computation}.
As such, we conduct back-substitution to compute the concrete bounds for neurons
in affine layer $\gamma_k$ (lines 11-27).

\begin{algorithm}[!ht]
  \textbf{Input:}
  $M$ is the network (eg. \autoref{fig:HypoNet});
  $\tau$ is the back-substitution threshold; $\segmpara$ is the network segmentation parameter\\

  \textbf{Annotatation:}
   input layer of $M$ as $\gamma_{in}$; constraint set of affine layer $\gamma_{k}$ as $\Upsilon_k$; the set of concrete bounds for neurons in layer $\gamma_{k} \in
  M$ as $C_k$; the \textit{segmented} network model as  $\mathcal{M}$\\

  \textbf{Assumption:}
  the analysis is conducted in ascending order of the layer index\\
  \textbf{Output:}
  tightest concrete bounds $C_k$ computed for all layer $\gamma_{k} \in M$

  \begin{algorithmic}[1]
  \STATE $\mathcal{M} \gets \Call{SegmentNetwork}{M, \segmpara}$
    \FORALL{layer $\gamma_k \in \mathcal{M}$}
    \IF{$\Call{IsReluLayer}{\gamma_k}$}
    \STATE  $\gamma_{pre} \gets \Call{PredecessorLayer}{\gamma_k}$
    \STATE $C_k \gets \Call{ComputeReluLayer}{\gamma_{pre}}$
    \ELSE
      \STATE $\gamma_{pre} \gets \Call{PredecessorLayer}{\gamma_k}$
      \STATE $\Upsilon_k \gets \Call{GetSymbolicConstraints}{\gamma_k}$
      \STATE $C_k \gets \Call{EvaluateConcreteBounds}{\Upsilon_k, \gamma_{pre}}$
      \STATE $counter_k =0$
      \WHILE{$\gamma_{pre} \neq  \gamma_{in}$}
        \IF{$\Call{IsEndLayer}{\gamma_{pre}}$}
          \STATE $sum \gets \Call{ReadSummary}{\gamma_{pre}}$
          \STATE $\Upsilon_k \gets \Call{BacksubWithBlockSummary}{\Upsilon_k, sum}$
          \STATE $counter_k \gets counter_k+1$
          \STATE $\gamma_{pre} \gets \Call{GetStartLayer}{\gamma_{pre}}$
        \ELSE
        \STATE $sym\_cons \gets \Call{GetSymbolicConstraints}{\gamma_{pre}}$
        \STATE $\Upsilon_k \gets \Call{BacksubWithSymbolicConstraints}{\Upsilon_k, sym\_cons}$
        \STATE $counter_k \gets counter_k+1$
        \STATE $\gamma_{pre} \gets \Call{PredecessorLayer}{\gamma_{pre}}$
      \ENDIF
      \IF{$\Call{IsEndLayer}{\gamma_k}$
          \AND $\gamma_{pre} = \Call{GetStartLayer}{\gamma_k}$}
        \STATE $\Call{StoreSummary}{\gamma_k, \Upsilon_k}$
      \ENDIF
      \STATE $temp\_ck \gets \Call{EvaluateConcreteBounds}{\Upsilon_k, \gamma_{pre}}$
      \STATE $C_k \gets \Call{UpdateBounds}{C_k, temp\_ck}$
       \IF{$counter_k \geq \tau$ \AND $\neg(\Call{IsEndLayer}{\gamma_k}$ \AND $\Call{LackSummary}{\gamma_k})$}
     \STATE \textbf{break}
    \ENDIF
    \ENDWHILE
    \ENDIF
  \ENDFOR
  \RETURN all $C_k$ for all layer $\gamma_k \in \mathcal{M}$
\end{algorithmic}
\caption{Overall analysis procedure in BBPoly}
\label{algo:methodalgo}
\end{algorithm}

We have two types of back-substitution and executing either one of the two will be considered as one step of back-substitution which leads to an increment of the counter for layer $\gamma_k$ (lines 15, 20):

\begin{itemize}[topsep=0pt, left=0em]
\item If  $\gamma_{pre}$ is the end layer of a block, we first read the block
  summary of $\gamma_{pre}$ (lines 12-13), and then call
  $\Call{BacksubWithBlockSummary}{\Upsilon_k, sum}$ to perform back-substituion
  over a block (line 14).
  After execution,  $\Upsilon_k$ will be updated to be defined over the start
  layer of the block. Lastly, in preparation for next iteration of execution,
  $\gamma_{pre}$ is set to the start layer of the block (line 16).

\item Otherwise, we conduct {\em layer-by-layer} back-substitution (lines 18-21)
  similarly to DeepPoly.
  We obtain $sym\_cons$, the symbolic constraints built for $\gamma_{pre}$, and
  call $\Call{BacksubWithSymbolicConstraints}{\Upsilon_k, sym\_cons}$ (line 19).
  Then, $\Upsilon_k$ will be updated to be defined over the predecessor layer of
  $\gamma_{pre}$.
  Pertaining to block summarization construction, if $\gamma_{pre}$ and
  $\gamma_k$ are the start and the end layer of the same block, $\Upsilon_k$
  will be recorded as the block summary (lines 22-23).
\end{itemize}

After generating a new set of constraints (lines 14, 19), we can compute a set
of potential concrete bounds $temp\_ck$ using the new constraints $\Upsilon_k$
defined over the new $\gamma_{pre}$ (line 24).
Then we update $C_k$ by the most precise bounds between $temp\_ck$ and the
previous $C_k$ (line 25) as proposed in DeepPoly, where \textit{the most
  precise} means the smallest upper bound and biggest lower bound.

\textbf{Bounded Back-substitution.}
Normally, we continue new constraint construction, constraint evaluation and
concrete bound update for $\gamma_k$ until the input layer (line 11).
The goal here is to explore the opportunity for cancellation of variables in the
constraints defined over a particular layer.
Such opportunity may lead to attaining tighter bounds for abstract values of
neurons at layer $k$.
Nevertheless, it is possible to terminate such back-substitution operation
earlier to save computational cost, at the risk of yielding less precise
results.\footnote{As a matter of fact, our empirical evaluation (detailed in
  Appendix~\ref{appendix:early_termi_appendix}) shows that the degree of
  improvement in accuracy degrades as we explore further back into earlier
  layers during back-substitution.}
This idea is similar in spirit to our introduction of block summarization. We
term such earlier termination as {\em bounded back-substitution.}
It may appear similar to the ``limiting back-substitution'' suggested in DeepPoly \cite{SinghGPV19popl} or GPUPoly \cite{ MullerSPT20}. However, we note that one step in back-substitution in our approach can either be a back-substitution over one layer or {\em over a block summary}.  
Please note the difference between our {\em bounded back-substitution} and the limiting back-substitution in DeepPoly \cite{SinghGPV19popl} or GPUPoly \cite{ MullerSPT20}. We count either a layer back-substitution or a back-substitution over block summary as one step of back-substitution. Therefore, even though we bound the same number of steps of back-substitution, our method allows us to obtain constraints defined over more preceding layers compared to limiting back-substitution in DeepPoly or GPUPoly.

Bounded back-substitution is incorporated in \autoref{algo:methodalgo}, by
accepting an input $\tau$, which is a threshold for the maximum number of steps
to be taken during back-substitution.
More specifically, we initialize a counter when processing layer $\gamma_k$
(line 10), and increment the counter accordingly during the analysis (lines 15,
20).
Finally, we end the back-substitution iteration for layer $\gamma_k$ once the
threshold is reached (line 26).

During the construction of block summarization, {\em we suspend this
  threshold checking when $\gamma_k$ is the end layer of a block} (second test
in line 26).
This ensures that the algorithm can generate its block summarization without
being forced to terminate early.
In summary, suppose each block has at most $\ell$ layers, under bounded
back-substitution, the layer $\gamma_k$ will back-substitute either $\ell$ layers
(if $\gamma_k$ is the end layer of a block) or $\tau$ steps (if $\gamma_k$ is
not the end layer of a block).

\subsection{Summarization within block}
\label{sec:summary_in_block}

\textbf{Block Summarization.}
\label{sec:formulation_blk_sum}
The summarization captures the relationship between the output neurons and input
neurons within a block.
Given a block with $k$ affine layers inside, we formally define it as $\Gamma
=\{\gamma_{\mathrm{in}}, \gamma_1, \gamma'_1,\dots, \gamma_k\}$ (e.g. block1 in
\autoref{fig:final_seg}) or $\Gamma =\{\gamma'_0, \gamma_1, \gamma'_1,\dots,
\gamma_k\}$ (like block2 in \autoref{fig:final_seg}), where $\gamma_i$ refers to
an affine layer, $\gamma_{\mathrm{in}}$ refers to the input layer and
$\gamma'_i$ refers to the ReLU layer with ReLU function applied on $\gamma_i$,
for $i\in\{0,1,2,\cdots, k\}$.

Suppose the last layer $\gamma_k = \{x_{k1}, \cdots, x_{kN}\}$
contains $N$ neurons in total.
The block summarization $\Phi_\Gamma = \{\langle \phi_{x_{k1}}^{L},
\phi_{x_{k1}}^{U} \rangle, \cdots,  \langle \phi_{x_{kN}}^{L}, \phi_{x_{kN}}^{U}
\rangle \}$ is defined as a set of constraint-pairs.
For $j \in \{1,2,\cdots, N\}$, each pair $\langle \phi_{x_{kj}}^{L},
\phi_{x_{kj}}^{U} \rangle$ corresponds to the lower and upper constraints of
neuron $x_{kj}$ defined over the neurons in the first layer of the block (be it
an affine layer $\gamma_{\mathrm{in}}$ or a ReLU layer $\gamma'_0$).
As these lower and upper constraints encode the relationship between output
neurons and input neurons with respect to the block $\Gamma$, they function as
the block summarization.

\textbf{Back-substitution with Block Summarization.}
\label{sec:formal_back_sub_with_blk_sum}
To explain our idea, we present the overall back-substitution process as the
matrix multiplication (cf. \cite{MullerSPT20}) depicted in
\autoref{fig:back_in_matrix}.
Matrix $M^k$ encodes the current constraints for neurons in layer $l$ defined
over neurons in previous layer $k$, where $1\leq k<l$.
The cell indexed by the pair $(x^l_{hm}, x^k_{ic})$ in the matrix records the
coefficient between neuron $x^l_{hm}$ in layer $l$ and neuron $x^k_{ic}$ in
layer $k$.
The same notation also applies for matrix $F^k$ and $M^{k-1}$, where $F^k$
denotes next-step back-substitution and $M^{k-1}$ represents a newly generated
constraint for neurons in layer $l$ defined over neurons in the preceding layer
$k-1$.
As we always over-approximate ReLU function to a linear function, without loss
of generality, we therefore discuss further by considering a network as a
composition of affine layers.

\begin{figure}[htbp]
\centering
    \begin{tikzpicture}
    \node[black] (origin){};
    \node[black, above left  = -0.5mm and -1 mm of origin] (){$x^l_{ht}$};
    \node[black, below right  = 0mm and 8 mm of origin] (){$M^k$};
    \node[black, above left  = 5.5mm and -1 mm of origin] (){$\cdots$};
    \node[black, above left  = 11.5mm and -1 mm of origin] (){$x^l_{h2}$};
    \node[black, above left  = 17.5mm and -1 mm of origin] (){$x^l_{h1}$};
     \node[black, above left  = 0mm and -6 mm of origin] (){$*$};
     \node[black, above left  = 23mm and -7 mm of origin] (){$x^k_{i1}$};
    \node[black, above left  = 5.5mm and -7 mm of origin] (){$\cdots$};
    \node[black, above left  = 11.5mm and -6 mm of origin] (){$*$};
    \node[black, above left  = 17.5mm and -6 mm of origin] (){$*$};
    \node[black, above left  = 0mm and -12 mm of origin] (){$*$};
    \node[black, above left  = 5.5mm and -13 mm of origin] (){$\cdots$};
    \node[black, above left  = 23mm and -13 mm of origin] (){$x^k_{i2}$};
    \node[black, above left  = 11.5mm and -12 mm of origin] (){$*$};
    \node[black, above left  = 17.5mm and -12 mm of origin] (){$*$};
   \node[black, above left  = 0mm and -19 mm of origin] (){$\cdots$};
   \node[black, above left  = 23mm and -19 mm of origin] (){$\cdots$};
    \node[black, above left  = 5.5mm and -19 mm of origin] (){$\cdots$};
    \node[black, above left  = 11.5mm and -19 mm of origin] (){$\cdots$};
    \node[black, above left  = 17.5mm and -19 mm of origin] (){$\cdots$};
   \node[black, above left  = 0mm and -24 mm of origin] (){$*$};
    \node[black, above left  = 5.5mm and -25 mm of origin] (){$\cdots$};
    \node[black, above left  = 23mm and -25 mm of origin] (){$x^k_{is}$};
    \node[black, above left  = 11.5mm and -24 mm of origin] (){$*$};
    \node[black, above left  = 17.5mm and -24 mm of origin] (){$*$};
    \node[black, right = 34mm of origin] (loc1){};
    \node[black, below right  = 0mm and 8 mm of loc1] (){$F^k$};
    \node[black, above left  = 9mm and 5 mm of loc1] (){$\circ$};
    \node[black, above left  = -0.5mm and -1 mm of loc1] (){$x^k_{is}$};
    \node[black, above left  = 5.5mm and -1 mm of loc1] (){$\cdots$};
    \node[black, above left  = 11.5mm and -1 mm of loc1] (){$x^k_{i2}$};
    \node[black, above left  = 17.5mm and -1 mm of loc1] (){$x^k_{i1}$};
     \node[black, above left  = 0mm and -6 mm of loc1] (){$*$};
     \node[black, above left  = 23mm and -8 mm of loc1] (){$x^{k-1}_{j1}$};
    \node[black, above left  = 5.5mm and -7 mm of loc1] (){$\cdots$};
    \node[black, above left  = 11.5mm and -6 mm of loc1] (){$*$};
    \node[black, above left  = 17.5mm and -6 mm of loc1] (){$*$};
    \node[black, above left  = 0mm and -12 mm of loc1] (){$*$};
    \node[black, above left  = 5.5mm and -13 mm of loc1] (){$\cdots$};
    \node[black, above left  = 23mm and -15 mm of loc1] (){$x^{k-1}_{j2}$};
    \node[black, above left  = 11.5mm and -12 mm of loc1] (){$*$};
    \node[black, above left  = 17.5mm and -12 mm of loc1] (){$*$};
   \node[black, above left  = 0mm and -19 mm of loc1] (){$\cdots$};
   \node[black, above left  = 23mm and -19 mm of loc1] (){$\cdots$};
    \node[black, above left  = 5.5mm and -19 mm of loc1] (){$\cdots$};
    \node[black, above left  = 11.5mm and -19 mm of loc1] (){$\cdots$};
    \node[black, above left  = 17.5mm and -19 mm of loc1] (){$\cdots$};
   \node[black, above left  = 0mm and -24 mm of loc1] (){$*$};
    \node[black, above left  = 5.5mm and -25 mm of loc1] (){$\cdots$};
    \node[black, above left  = 23mm and -26 mm of loc1] (){$x^{k-1}_{jr}$};
    \node[black, above left  = 11.5mm and -24 mm of loc1] (){$*$};
    \node[black, above left  = 17.5mm and -24 mm of loc1] (){$*$};
    \node[black, right = 33.5mm of loc1] (loc2){};
    \node[black, below right  = 0mm and 6 mm of loc2] (){$M^{k-1}$};
    \node[black, above left  = 9mm and 5 mm of loc2] (){$=$};
    \node[black, above left  = -0.5mm and -1 mm of loc2] (){$x^l_{ht}$};
    \node[black, above left  = 5.5mm and -1 mm of loc2] (){$\cdots$};
    \node[black, above left  = 11.5mm and -1 mm of loc2] (){$x^l_{h2}$};
    \node[black, above left  = 17.5mm and -1 mm of loc2] (){$x^l_{h1}$};
     \node[black, above left  = 0mm and -6 mm of loc2] (){$*$};
     \node[black, above left  = 23mm and -8 mm of loc2] (){$x^{k-1}_{j1}$};
    \node[black, above left  = 5.5mm and -7 mm of loc2] (){$\cdots$};
    \node[black, above left  = 11.5mm and -6 mm of loc2] (){$*$};
    \node[black, above left  = 17.5mm and -6 mm of loc2] (){$*$};
    \node[black, above left  = 0mm and -12 mm of loc2] (){$*$};
    \node[black, above left  = 5.5mm and -13 mm of loc2] (){$\cdots$};
    \node[black, above left  = 23mm and -15 mm of loc2] (){$x^{k-1}_{j2}$};
    \node[black, above left  = 11.5mm and -12 mm of loc2] (){$*$};
    \node[black, above left  = 17.5mm and -12 mm of loc2] (){$*$};
   \node[black, above left  = 0mm and -19 mm of loc2] (){$\cdots$};
   \node[black, above left  = 23mm and -19 mm of loc2] (){$\cdots$};
    \node[black, above left  = 5.5mm and -19 mm of loc2] (){$\cdots$};
    \node[black, above left  = 11.5mm and -19 mm of loc2] (){$\cdots$};
    \node[black, above left  = 17.5mm and -19 mm of loc2] (){$\cdots$};
   \node[black, above left  = 0mm and -24 mm of loc2] (){$*$};
    \node[black, above left  = 5.5mm and -25 mm of loc2] (){$\cdots$};
    \node[black, above left  = 23mm and -26 mm of loc2] (){$x^{k-1}_{jr}$};
    \node[black, above left  = 11.5mm and -24 mm of loc2] (){$*$};
    \node[black, above left  = 17.5mm and -24 mm of loc2] (){$*$};
   \draw [step=0.6,black, thin] (0,0) grid (2.4,2.4);
   \draw [step=0.6,black, thin] (3.6,0) grid (6,2.4);
   \draw [step=0.6,black, thin] (7.2,0) grid (9.6,2.4);
    \end{tikzpicture}
\caption{Back-substitution process can be represented as matrix multiplication with constant terms (e.g. biases) being omitted, cf. \cite{MullerSPT20} }
\label{fig:back_in_matrix}
\end{figure}

Next, we describe how to perform back-substitution with the generated block
summarization.
After completing the layer-by-layer back-substitution process within a given
block (take block 2 in \autoref{fig:final_seg} as example), we obtain
constraints of neurons in the affine layer 6 ($\gamma_6$) defined over neurons
in the ReLU layer 3 ($\gamma'_3$), which corresponds to $M^k$.
This matrix is then multiplied with matrix $F^{k1}$ which captures the affine
relationship between neurons in layer $\gamma_3^\prime$ and $\gamma_3$ (this
affine relationship is actually an over-approximation since
$\gamma_3^\prime=\mathrm{ReLU}(\gamma_3))$, followed by another multiplication
with matrix $F^{k2}$ constructed from block summarization for block 1 (in
\autoref{fig:final_seg}), denoted here by $\Phi_{\Gamma_1}$.
$\Phi_{\Gamma_1}$ is a set of constraints for neurons in the affine layer 3
($\gamma_3$) defined over neurons in the input layer ($\gamma_{in}$) and is
computed already during the analysis of block 1.
Hence, the resulting matrix $M^k \circ F^{k1} \circ F^{k2}$ encodes the
coefficients of neurons in layer $\gamma_6$ defined over neurons in layer
$\gamma_\mathrm{in}$.
Through this process, we achieve back-substitution of the constraints of layer
$\gamma_6$ to the input layer.

\textbf{Memory Usage and Time Complexity.}
\label{sec:blk_sum_momery_requriement}
In the original method, the memory usage of DeepPoly is high since it associates all
neurons with symbolic constraints and maintains all symbolic constraints
throughout the analysis process for the sake of layer-by-layer
back-substitution.
In work of \cite{BalutaCMS21} and \cite{TranBXJ20}, they all faced with
out-of-memory problem when running DeepPoly on their evaluation platform.
In our block summarization approach, a block captures only the relationship
between its end and start layers.
Consequently, all the symbolic constraints for intermediate layers within the
block can be released early once we complete the block summarization computation
(illustrated by a delete icon in the layer in \autoref{fig:net_in_blks}).
Thus our method requires less memory consumption when analyzing the same
network, and the memory usage can also be controlled using the network
segmentation parameter $\segmpara$.

For time complexity, consider a network with $n$ affine layers and each layer
has at most $N$ neurons, DeepPoly's complexity is $O(n^2\cdot N^3)$.
In our method, with bounded back-substitution (detail in
\autoref{sec:form_modul}), we can bound the number of steps for
back-substitution to a constant for each layer.
Thus the time complexity can be reduced to $O(n\cdot N^3)$.
Without bounded back-substitution, we have constant-factor reduction in time
complexity, yielding the same $O(n^2\cdot N^3)$.

\vspace{-0.8em}
\subsection{Summarization defined over input layer}
\label{sec:summaryoverinput}

Previously, \autoref{sec:summary_in_block} describes a back-substitution
mechanism on ``block-by-block'' basis.
To further simplify the back-substitution process and save even more on the
execution time and memory, we also design a variation of block summarization
that is {\em defined over the input layer}.
As the overall procedure of back-substitution with summarization defined over
input layer is similar to the block summarization described in
\autoref{algo:methodalgo}, we provide the algorithm for this new summary in
Appendix \ref{appendix:algo-sum-input}.

\textbf{Summary over Input.}
Just as in \autoref{sec:formulation_blk_sum}, the summary-over-input is still
formulated as $\Phi_\Gamma$.
However, $\langle \phi_{x_{jk}}^{L}, \phi_{x_{jk}}^{U} \rangle$ corresponds to
constraints of neuron $x_{jk}$ which are now {\em defined over the input
  neurons}.
To generate summary for block $\Gamma_i$, we firstly do layer-by-layer analysis
within the block, then we back-substitute further with the summary for block
$\Gamma_{i-1}$ which is defined over input neurons, thus we get
$\Phi_{\Gamma_i}$ defined over the input neurons.

\textbf{Back-substitution with Summary over Input.}
The back-substitution process also follows the formulation described in \autoref{sec:formal_back_sub_with_blk_sum}.
The resulting matrix $M^k \circ F^{k1} \circ F^{k2}$ will directly be defined
over input neurons since $F^{k2}$ is the summary of preceding block directly
defined over input neurons.

\textbf{Memory Requirement and Time Complexity.}
Once the summary generation for block $\Gamma_i$ has completed, all symbolic
constraints and summaries from previous $i-1$ blocks could be released, only
the input layer needs to be kept.
For time complexity, each layer back-substitutes at most $l+1$ steps (if each
block has maximum $l$ layers), the time complexity will be $O(n\cdot N^3)$.

\vspace{-1em}

\section{Experiment}
\vspace{-0.5em}

We have implemented our proposed method in a prototype analyzer called BBPoly,
which is built on top of the state-of-the-art verifier DeepPoly.
Then, we conducted extensive experiments to evaluate the performance of both our
tool and DeepPoly, in terms of precision, memory usage and runtime.
In the following subsections, we will describe the details of our experiment.

\vspace{-1em}
\subsection{Experiment Setup}

We propose two types of block summary in our BBPoly system:

\begin{itemize}[noitemsep, topsep=0pt]
\item Block summarization as described in \autoref{sec:summary_in_block}. It can
  be supplemented with bounded back-substitution heuristic in
  \autoref{sec:form_modul} to facilitate the analysis even more for extremely
  large network;

\item Block summary defined over input layer that is introduced in
  \autoref{sec:summaryoverinput}
\end{itemize}

We compare our methods with the state-of-the-art system DeepPoly
\cite{SinghGPV19popl}, on top of which our prototype tool is built.
DeepPoly is publicly available at the GitHub repository of the ERAN system \cite{ERANSystem}. 
On the other hand, we do not compare with existing approach using MILP solving \cite{DBLP:conf/aaai/BotoevaKKLM20} since the latter can only handle small networks, such as MNIST/CIFAR10 networks with 2 or 3 hidden layers while our BBPoly can analyze large networks of up to 34 hidden layers.

\textbf{Evaluation datasets.}
We choose the popular MNIST \cite{lecun-mnisthandwrittendigit-2010} and CIFAR10
\cite{cifar10dataset} image datasets that are commonly used for robustness
verification.
MNIST contains gray-scale images with $28 \times 28$ pixels and CIFAR10 consists
of RGB 3-channel images of size $32 \times 32$.
Our test images are provided from DeepPoly paper where they select out the first
100 images of the test set of each dataset.
The test images are also publicly available at \cite{ERANSystem}.

\textbf{Evaluation platform and networks.}
The evaluation machine is equipped with a 2600 MHz 24 core GenuineIntel CPU with
64 GB of RAM.
We conducted experiments on networks of various sizes as itemized in
\autoref{tab:net_for_precision}; these include fully-connected, convolutional
and (large sized) residual networks where the number of hidden neurons is up to
967K.
All networks use ReLU activation, and we list the layer number and number of
hidden neurons in the table.
Specifically, the networks whose names suffixed by ``DiffAI'' were trained with
adversarial training DiffAI \cite{MirmanGV18}.
These benchmarks are also collected from \cite{ERANSystem}.

\textbf{Verified robustness property.}
We verify robustness property against the $L_\infty$ norm attack
\cite{Carlini017} which is paramterized by a constant $\epsilon$ of
perturbation.
Originally, each pixel in an input image has a value $p_i$ indicating its color
intensity.
After applying the $L_\infty$ norm attack with a certain value of $\epsilon$,
each pixel now corresponds to an intensity interval $[p_i-\epsilon,
  p_i+\epsilon]$.
Therefore we constructed an adversarial region defined as $\varprod_{i=1}^n
[p_i-\epsilon, p_i+\epsilon]$.
Our job is to verify that whether a given neural network can classify all
perturbed images within the given adversarial region as the same label as
of the original input image.
If so, we conclude that robustness is verified for this input image, the given
perturbation $\epsilon$ and the tested network.
For images that fail the verification, due to the over-approximation error, we
fail to know if the robustness actually holds or not, thus we report that the
results are inconclusive.
We set a 3-hour timeout for the analysis of each image, if the verifier fails to
return the result within 3 hours, we also state that the result is inconclusive.

\begin{table}[ht]
  \caption{Experimental Networks}
  \centering
  \def\arraystretch{1.1}
  \begin{tabular}{|l|c|c|c|c|c|}
    \hline
    \textbf{Neural Network} &
    \textbf{Dataset} &
    \textbf{\#Layer} &
    \textbf{\#Neurons} &
    \textbf{Type} & \textbf{Candidates}\\

    \hline
    MNIST\_9\_200  &   MNIST & 9 & 1,610& fully connected & 97\\
    \hline
    ffcnRELU\_Point\_6\_500 & MNIST & 6 & 3,000 & fully connected & 100 \\
    \hline
    convBigRELU  & MNIST & 6 & 48,064 & convolutional & 95\\
    \hline
    convSuperRELU  & MNIST & 6 & 88,544 & convolutional & 99\\
    \hline
    ffcnRELU\_Point\_6\_500  & CIFAR10 & 6 & 3,000 & fully connected & 56\\
    \hline
    convBigRELU & CIFAR10 \ & 6 & 62,464 & convolutional & 60\\
    \hline
    SkipNet18\_DiffAI  & CIFAR10 & 18 & 558K & residual & 41\\
    \hline
    ResNet18\_DiffAI  & CIFAR10 & 18 & 558K & residual & 46\\
    \hline
    ResNet34\_DiffAI  & CIFAR10 & 34 & 967K & residual & 39\\
    \hline
  \end{tabular}
  \label{tab:net_for_precision}
\end{table}
\vspace{-1em}
\subsection{Experiments on fully-connected and convolutional networks}
\label{sec:Exp-full-conv}

We firstly present the experiment results on fully-connected and convolutional
networks for both the MNIST and CIFAR10 datasets.
We set the block segmentation parameter to be 3;\footnote{We have conducted preliminary experiments with the effectiveness of having different block sizes; the results are available in Appendix \ref{appendix:diff-blk}. A more thorough investigation on the effectiveness is left as future work.} this means there will be 3
affine layers contained in a block.
We conduct experiments on both block-summarization and summary-over-input
methods.
And the bounded back-substitution heuristic is disabled for this part of
experiments.
We set six different values of perturbation $\epsilon$ for different networks
according to the settings in DeepPoly (details in Appendix
\ref{appendix:perturbation-size}).

\begin{figure*}[!ht]
  \begin{minipage}{0.95\linewidth}
    \begin{subfigure}[b]{0.48\textwidth}
      \centering
      \includegraphics[width=\textwidth]{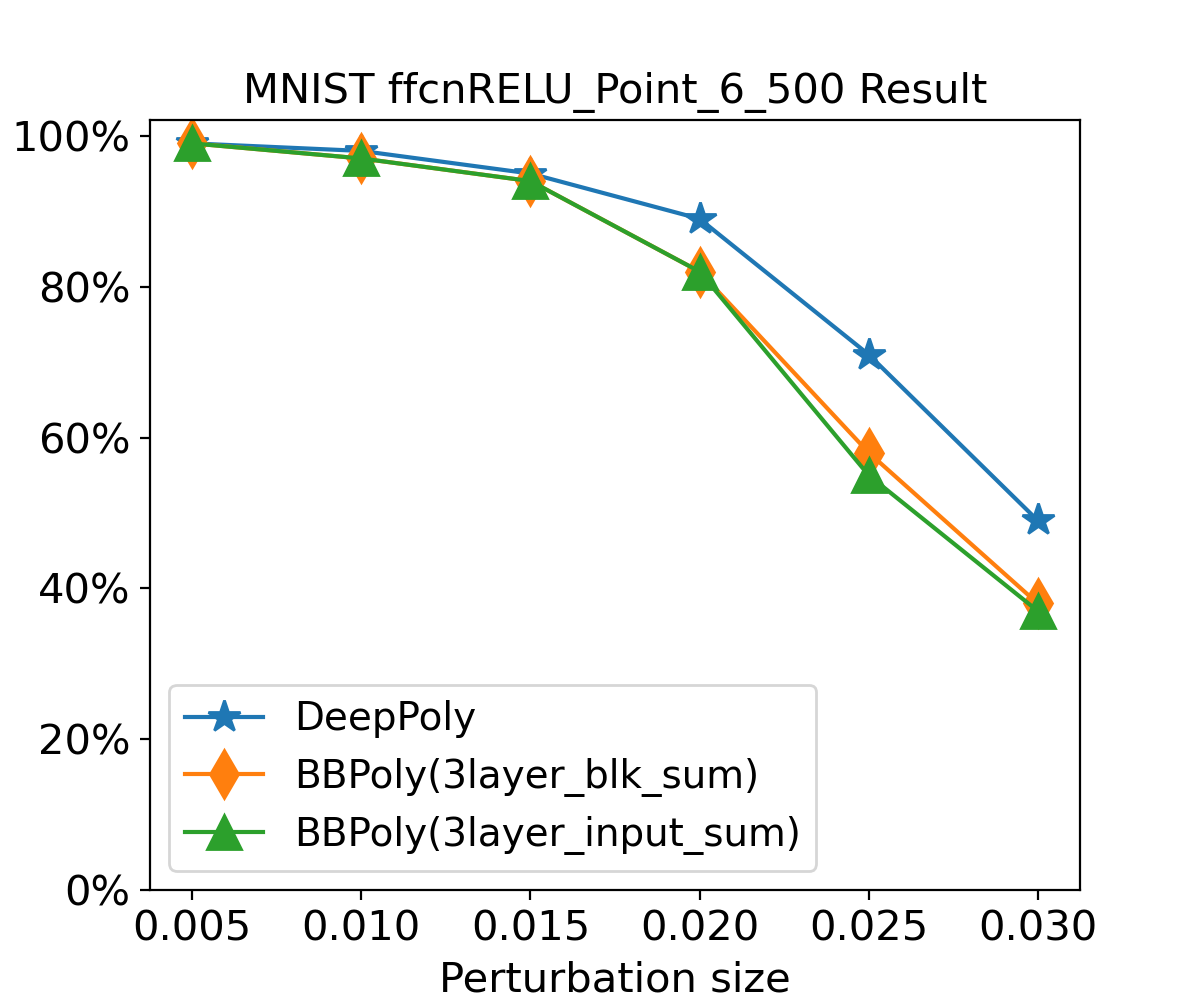}
      \vspace{-1.5em}
      \caption{}%
    \end{subfigure}
    \hfill
    \begin{subfigure}[b]{0.48\textwidth}
      \centering
       \includegraphics[width=\textwidth]{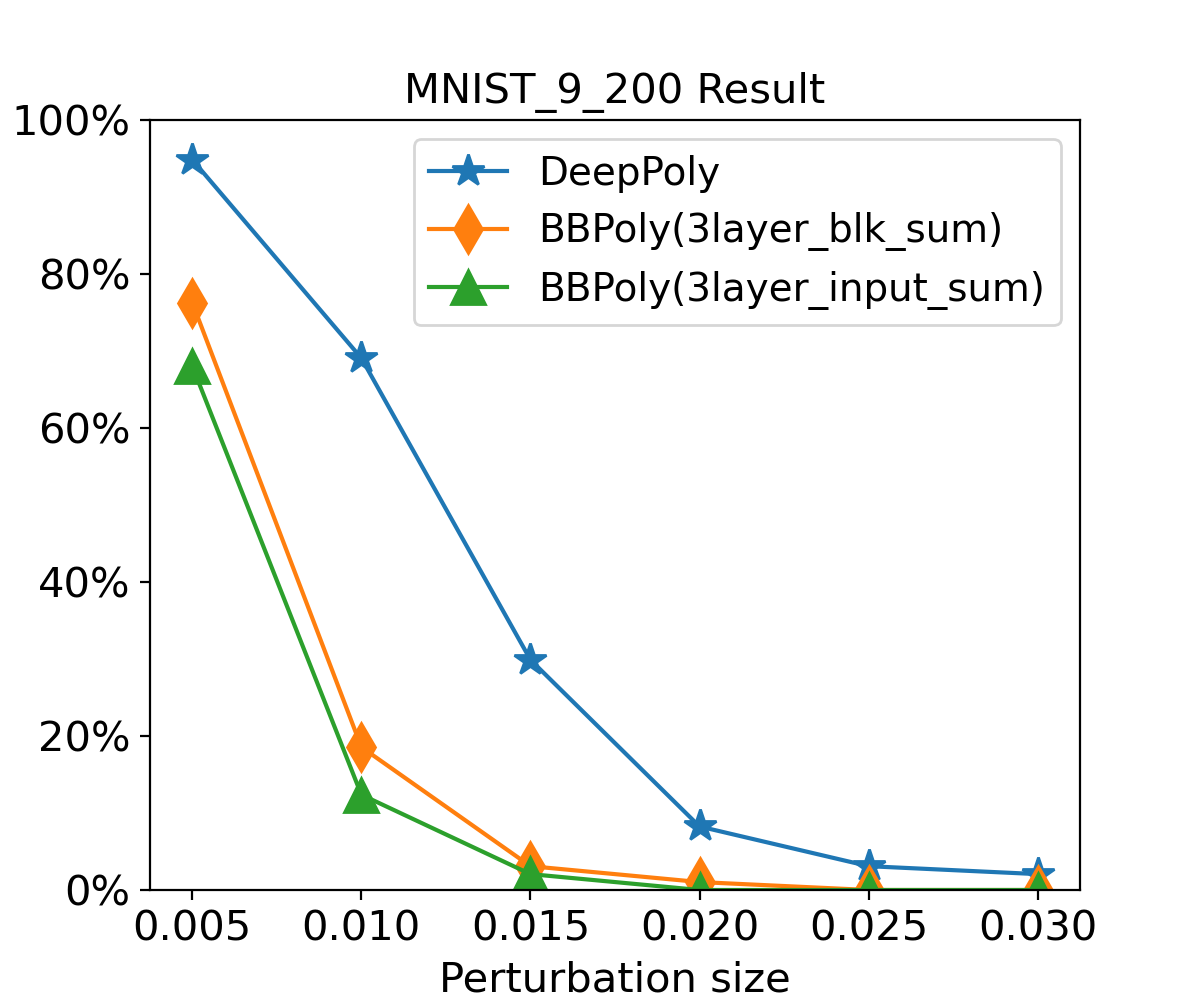}
      \vspace{-1.5em}
      \caption{}%
      \label{fig:pre-mnist-9-200}
    \end{subfigure}
    \vspace{-0.5em}
  \end{minipage}

  \begin{minipage}{0.95\linewidth}
    \begin{subfigure}[b]{0.48\textwidth}
      \centering
      \includegraphics[width=\textwidth]{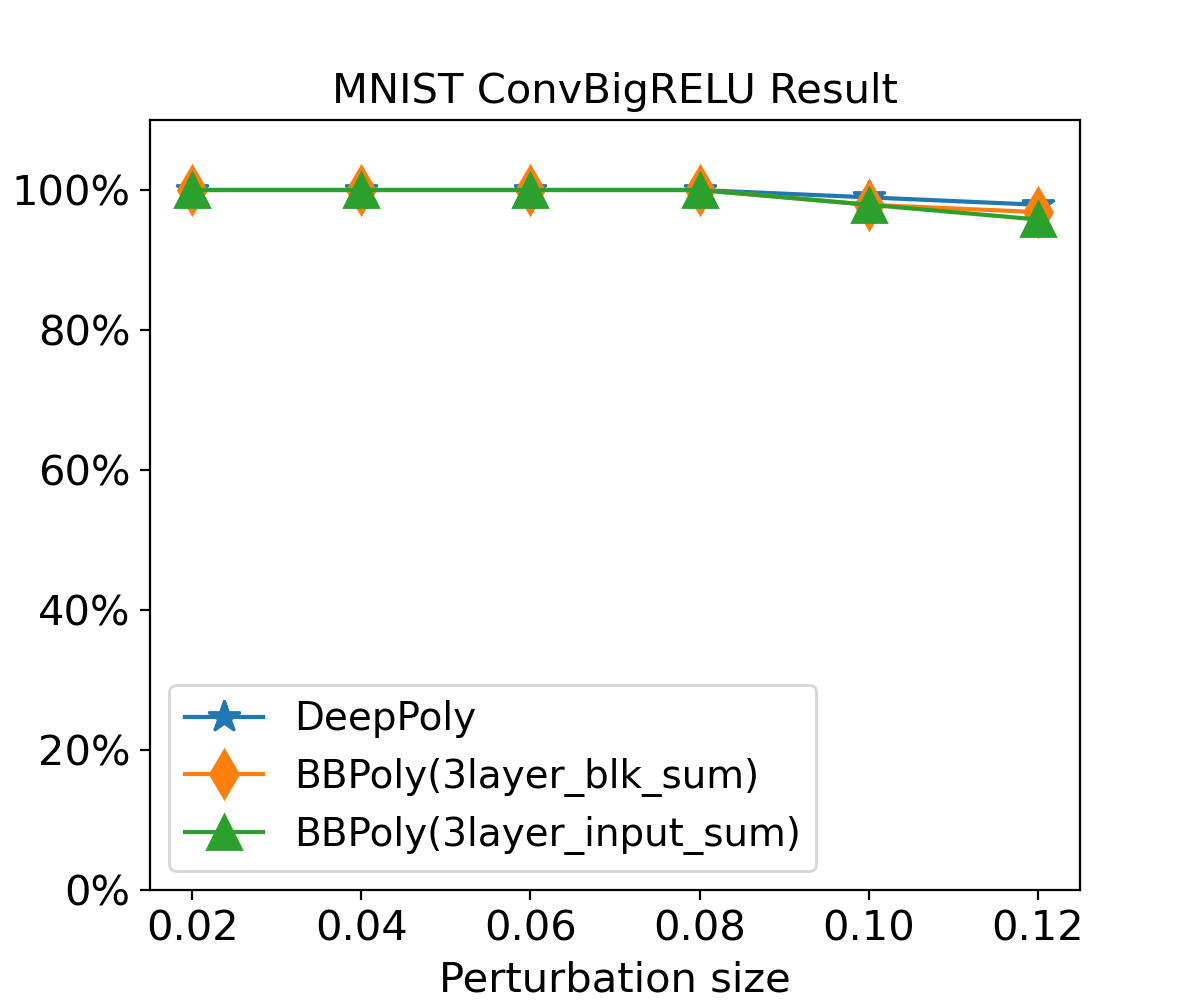}
      \vspace{-1.5em}
      \caption{}%
      \label{fig:mnist_convbig_prec}
    \end{subfigure}
    \hfill
    \begin{subfigure}[b]{0.48\textwidth}
      \centering
      \includegraphics[width=\textwidth]{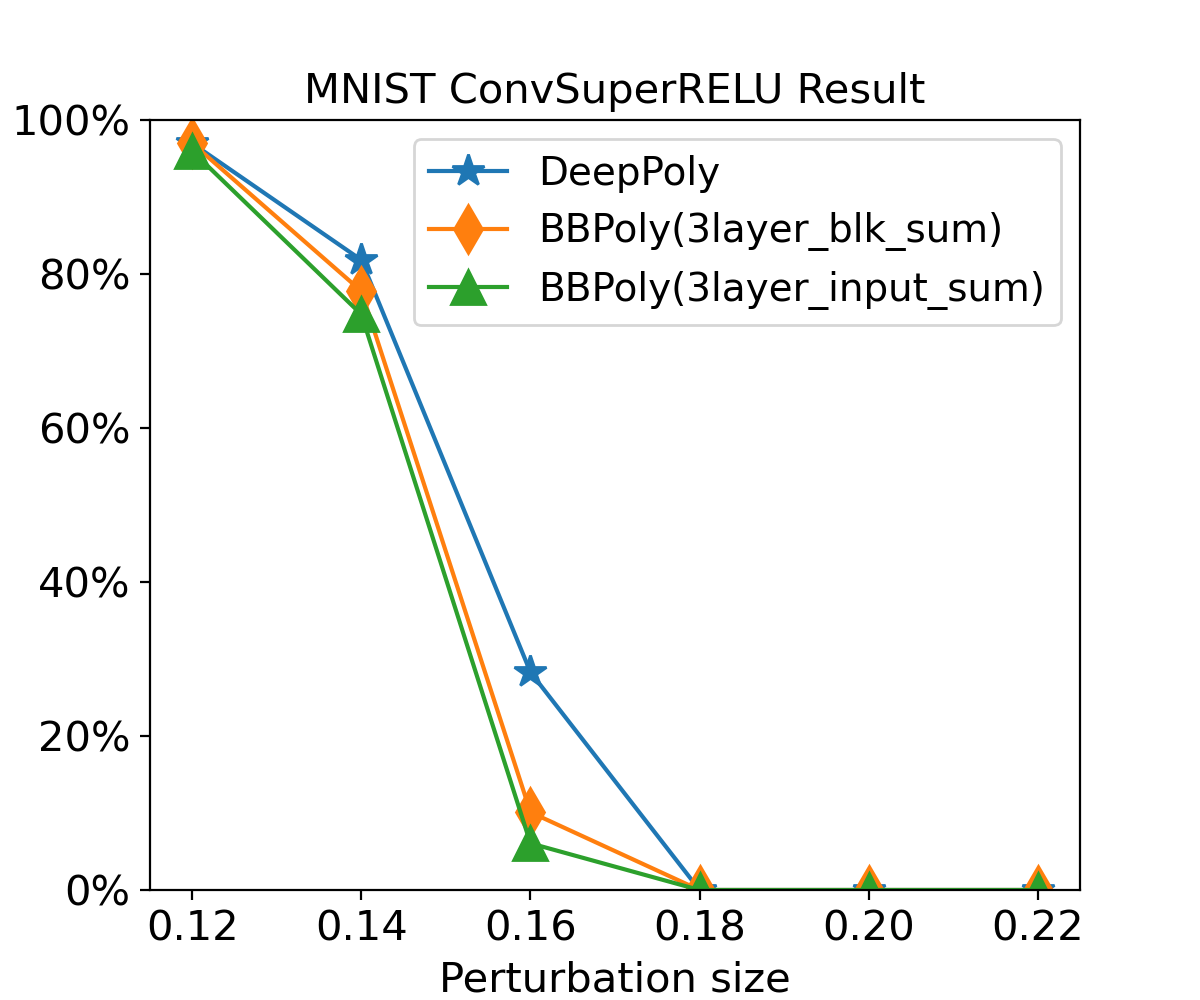}
      \vspace{-1.5em}
      \caption{}%
    \end{subfigure}
  \end{minipage}

  \vspace{-0.1em}
  \caption{Verified robustness precision comparison between our BBPoly system
    and DeepPoly for MNIST fully-connected and convolutional networks}
    \vspace{-0.7em}
  \label{fig:mnist_ffcn_conv_precision}
\end{figure*}

\begin{figure*}[!ht]
  \begin{minipage}{1\linewidth}
    \hspace{-1em}
    \begin{subfigure}[b]{0.48\textwidth}
      \includegraphics[width=\linewidth]{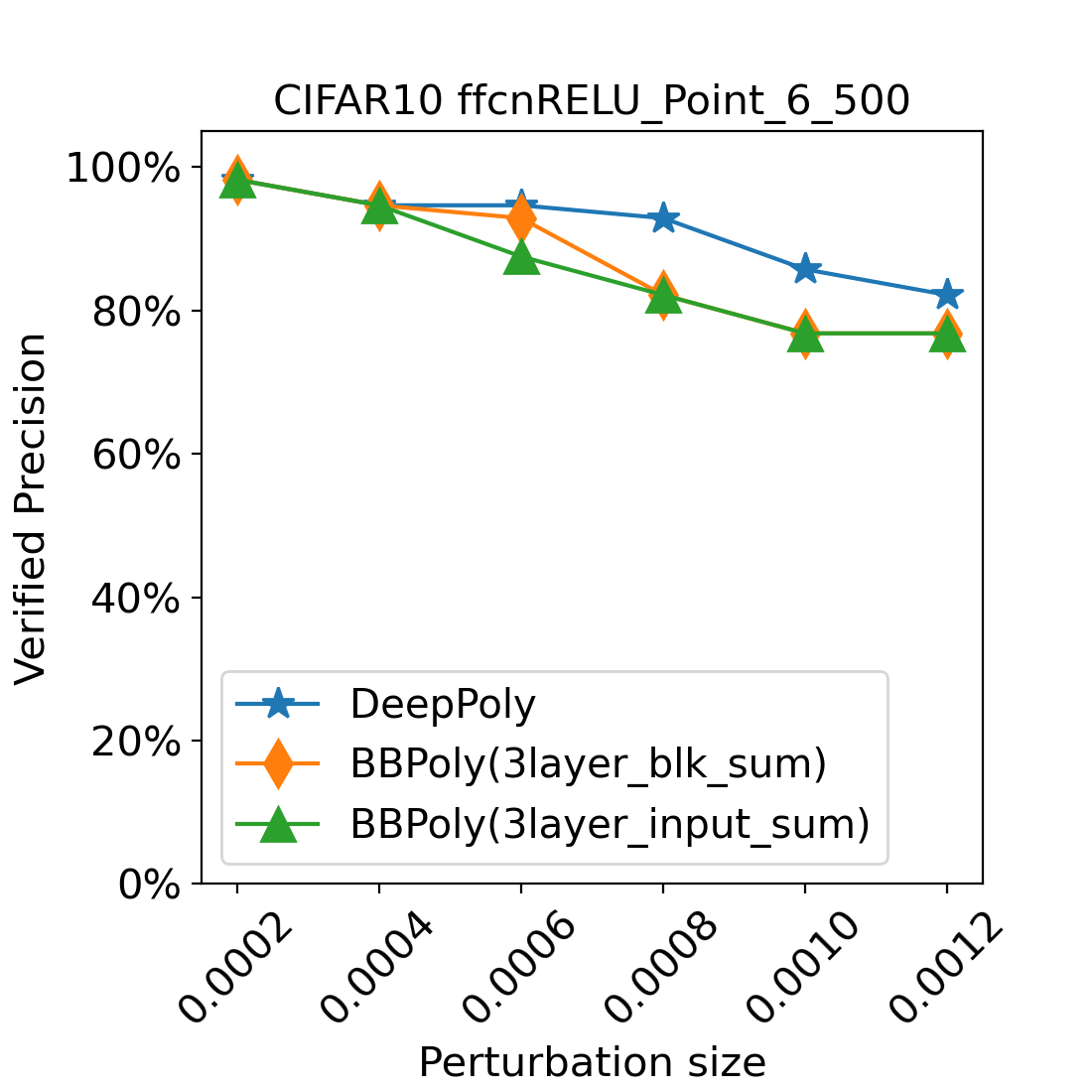}
      \caption{}%
    \end{subfigure}
    \hspace{-1.5em}
    \begin{subfigure}[b]{0.48\textwidth}
      \includegraphics[width=\linewidth]{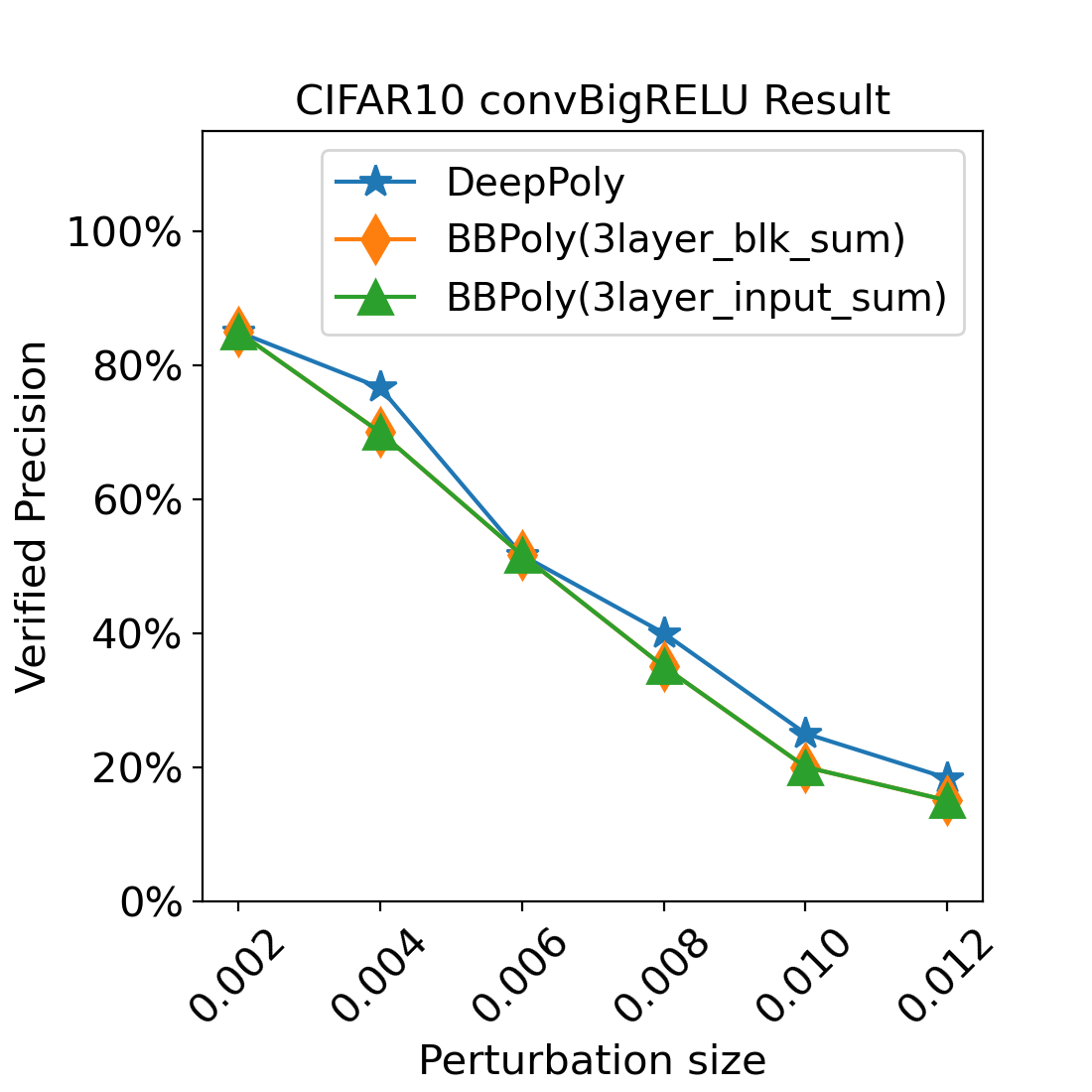}
      \caption{}%
    \end{subfigure}
    \caption{Verified robustness precision comparison  between our BBPoly system and DeepPoly for CIFAR10
      fully-connected and convolutional networks.}
    \label{fig:cifar_ffcn_conv_precision}
  \end{minipage}
\end{figure*}

The verified precision is computed as follows:

\begin{equation}
\begin{array}[t]{c}
\textrm{Number of verified images} \\ \hline
\textrm{Number of candidate images}
\end{array}
\label{precision-formula}
\end{equation}
where candidate images are those which have been correctly classified by a network. The numbers of candidate images for each network are presented in \autoref{tab:net_for_precision}.
\autoref{fig:mnist_ffcn_conv_precision} shows the precision comparison among
different methods on MNIST networks, and \autoref{fig:cifar_ffcn_conv_precision}
shows the precision on CIFAR10 networks.
Due to page constraint, full details of the precision and average execution time
per image for the experiments are recorded in \autoref{tab:precision_conv_ffcn}
of Appendix \ref{appendix:precision_in_tables}.
As expected, $\mathrm{DeepPoly} \geq \mathrm{BBPoly~(using~block~summary)} \geq
\mathrm{BBPoly~(using~input~summary)}$ with respect to precision and execution
time.
Apart from MNIST\_9\_200 network, our methods actually achieve comparable
precision with DeepPoly.

With regard to the execution time, for larger networks such as the three
convolutional networks experimented, our block-summarization method can save
around half of the execution time in comparison with that by DeepPoly.
Interested reader may wish to refer to \autoref{tab:precision_conv_ffcn} in
Appendix \ref{appendix:precision_in_tables} for detail.
The execution time can be significantly reduced for even larger network, such as
the deep residual networks, as demonstrated in \autoref{sec:expr_residual_nets}.

\vspace{-1em}
\vspace{-0.8em}

\subsection{Experiments on residual networks}
\label{sec:expr_residual_nets}

\textbf{Network description.}
We select three residual networks that have 18 or 34 layers and contain up to
almost one million neurons as displayed in \autoref{tab:net_for_precision}.
The SkipNet18, ResNet18 and ResNet34 are all trained with DiffAI defence.

\textbf{Perturbation size.}
DeepPoly is not originally designed to handle such large networks and is
inconclusive within our timeout.
However, an efficient GPU implementation of DeepPoly (called GPUPoly)
\cite{MullerSPT20} is proposed for much larger networks.
GPUPoly achieves the same precision as DeepPoly and it selects $\epsilon =
8/255$ for our experimental residual networks.
Thus we follow the same setting as in GPUPoly.
Unfortunately, GPUPoly does not run in one-CPU environment, and thus not
comparable with our experimental setting.

\textbf{Precision comparison with DeepPoly.}
\label{sec:prec_comp_with_gpupoly}
We only conduct robustness verification on candidate images as in \autoref{sec:Exp-full-conv}.
We set our baselines to be block-summarization method with bounded
back-substitution in four steps, and summary-over-input method.
The number of candidate images, verified images and the average execution time
per image for our experiment are listed in \autoref{tab:precision_residual},
where column ``BlkSum\_4bound'' refers to block-summarization method together
with bounded back-substitution in four steps and ``Input\_Sum'' refers to our
summary-over-input method.
As illustrated in \autoref{tab:precision_residual}, the summary-over-input
method verifies more or at least the same number of images compare to
the block-summarization method with bounded back-substitution but requires less
execution time, which demonstrates the competitive advantage of
our summary-over-input method.

\begin{table}[!ht]
  \centering
  \caption{The number of verified images and average execution time per image for
    CIFAR10 residual networks}
  \vspace{0.5em}
  \def\arraystretch{1.2}
  \begin{tabular}{|c|c|c|c|c|c|c|c|c|}
    \hline
    \multirow{2}{*}{\textbf{Neural Net}} &
    \multirow{2}{*}{$\epsilon$} &
    \multirow{2}{*}{
      \makecell{\textbf{Cand-} \\[-0.2em] \textbf{idates}}} &
    \multicolumn{2}{c|}{
      \makecell{\textbf{BBPoly} \\[-0.2em] {{\footnotesize (BlkSum\_4bound)}}}} &
    \multicolumn{2}{c|}{
      \makecell{\textbf{BBPoly} \\[-0.2em] {{\footnotesize (Input\_Sum)}} }} &
    \multicolumn{2}{c|}{\textbf{DeepPoly}}\\

    \cline{4-9}
    & & &
    Verified & Time(s) &
    Verified & Time(s) &
    Verified & Time(s) \\
    \hline

    SkipNet18\_DiffAI & 8/255 & 41 & 35 & 4027.08 & 36 & 1742.93 & - & -\\
    \hline
    ResNet18\_DiffAI & 8/255 & 46 & 29 & 3212.26 & 29 & 984.43 & - & -\\
    \hline
    ResNet34\_DiffAI & 8/255 & 39 & 21 & 2504.89 & 22 & 1296.78 & - & -\\
    \hline
  \end{tabular}
  \vspace{-2em}
  \label{tab:precision_residual}
\end{table}

\begin{table}[ht]
  \caption{Verified precision comparison computed from
    \autoref{tab:precision_residual}}
  \vspace{0.5em}
  \def\arraystretch{1.2}
  \centering
  \begin{tabular}{|c|c|c|c|c|}
    \hline
    \textbf{Neural Net} &
    \textbf{$\epsilon$}  &
    \makecell{\textbf{BBPoly} \\[-0.2em] {{\footnotesize (BlkSum\_4bound)}}} &
    \makecell{\textbf{BBPoly} \\[-0.2em] {{\footnotesize (Input\_Sum)}}} &
    \textbf{DeepPoly} \\

    \hline
    SkipNet18\_DiffAI & 8/255 & 85.3\% & 87.8\% & -\\
    \hline
    ResNet18\_DiffAI & 8/255 & 63.0\% & 63.0\%  & -\\
    \hline
    ResNet34\_DiffAI & 8/255 & 53.8\% & 56.4\%  & - \\
    \hline
  \end{tabular}
  \label{tab:resnet_precision}
\end{table}

Verified precision is computed using formula \ref{precision-formula} with data from
\autoref{tab:precision_residual}; the results are displayed in \autoref{tab:resnet_precision}
for residual networks.
DeepPoly fails to verify any image within the timeout of 3 hours in our
evaluation platform (indicated by `-') whereas our method yields reasonable verified precision within this time limit, supporting our hypothesis that BBPoly can scale up to analyze large networks with fair execution time and competitive precision.

\textbf{Time comparison with DeepPoly.}
To the best of our knowledge, there is no public experimental result of using
DeepPoly to analyze ResNets.
We initially used DeepPoly to analyze input images in our dataset with a smaller
$\epsilon = 0.002$ for ResNet18\_DiffAI.
Since DeepPoly took around 29 hours to complete the verification of an image, we
could not afford to run DeepPoly for all 100 test images.
In contrast, our summary-over-input method takes only 1319.66 seconds ($\approx$ 22
minutes) for the same image.
We also try to analyze ResNet18\_DiffAI with $\epsilon = 8/255$ according to the
original perturbation setting, and DeepPoly takes around 41 hours to complete
the verification of one image.
On the other hand, our block-summarization with bounded back-substitution in 4
steps uses average 3212.26 seconds ($\approx$ 54 minutes) for one image.

\textbf{Memory comparison with DeepPoly.}
We mention earlier that our methods utilize less memory.
To empirically testify this, we compare the peak memory usage between DeepPoly
and summary-over-input method with respect to ResNet18\_DiffAI, on the first
input image in our dataset and $\epsilon=8/255$.
We use the following command to check the peak memory usage of our analysis process:

\centerline{\texttt{\$ grep VmPeak /proc/\$PID/status}}

According to the result, DeepPoly takes up to 20.6 GB of memory while our
summary-over-input method needs much less memory.
It takes only 11.4 GB of memory, which is 55\% of the memory usage of DeepPoly.

\vspace{-1em}

\section{Discussion}
\vspace{-0.5em}

We now discuss the limitation of our work.
There are two limitations as follows.
Firstly, although the experimental results in \autoref{sec:Exp-full-conv}
demonstrate that our tool yields comparable precision with DeepPoly for majority
of the tested networks, it still significantly less precise than DeepPoly in
certain benchmarks, such as the MNIST\_9\_200 network.
We have explained earlier that this loss of precision is due to our current
block summarization technique which cannot capture a precise enough relationship
between neurons in the start and the end layer of a block.
In the future, we aim to generate a more tightened summarization to reduce the
over-approximation error and increase the precision of our analyzer.
Secondly, our current construction of a block is simple and straightforward.
We currently fix the block size to be a constant (eg. 3), and have not considered the intricate
information related to the connectivity between layers when choosing the block
size.
For future work, we will investigate how to utilize such information to assign
the block size dynamically.
This could potentially help the analysis to find a better trade-off between
speed and precision.

Our proposed method on block summarization could potentially be applied to other
neural network verification techniques to enhance their scalability.
For instance, in constraint-based verification, the network is formulated by the
conjunction of the encoding of all neurons and all connections between neurons
\cite{albarghouthi-book}.
This heavy encoding is exact but lays a huge burden on the constraint solver.
Following our block summary method, we could generate over-approximate encoding
of the network block to summarize the relationship between the start layer and
end layer of the block.
This could potentially lead to a significant decrease in the number of
constraints and make such analyzer more amenable to handle larger networks.

\vspace{-1em}


\section{Related Work}
\vspace{-0.5em}
Existing works on analyzing and verifying the robustness of neural networks can
be broadly categorized as \emph{complete} or \emph{incomplete} methods: given
sufficient time and computational resource, a complete verifier can always
provide a definite answer (\emph{yes} or \emph{no}) indicating whether a neural
network model is robust or not, while an incomplete verifier might return an
\emph{unknown} answer.

Typical complete methods include the works in \cite{TjengXT19, BotoevaKKLM20,
  KatzBDJK17, KatzHIJLLSTWZDK19}, which encode the verification problems into
arithmetic constraints, and utilize the corresponding sound and complete solvers
to solve them.
In particular, the techniques in \cite{TjengXT19, BotoevaKKLM20} are based on
MILP (mixed integer liner program) solvers, while the verifiers in
\cite{KatzBDJK17, KatzHIJLLSTWZDK19} use SMT (satisfiability modulo theory)
solvers in the theory of linear real arithmetic with ReLU constraints.
Although these solvers can give precise answers, they are also costly when
handling a large set of constraints with many variables.
Hence, it is difficult for complete verifiers to scale up.

In a different approach, the works \cite{PulinaT10, GehrMDTCV18, SinghGPV19popl}
introduce incomplete methods which over-approximate the behavior of neural
networks using techniques like abstraction interpretation, reachability analysis
etc.
Even though they might lose precision in certain situations, they are more
scalable than those complete methods.
The abstract domain devised for abstract interpretation is the
essential part of the analysis.
There has been progress in the design of abstract domains,
from interval domains in \cite{PulinaT10} to zonotope domains in
\cite{GehrMDTCV18} and finally to polyhedral domains in \cite{SinghGPV19popl}.
These domains allow the verifiers to prove more expressive specifications, such
as the robustness of neural networks, and handle more complex networks, like the
deep convolutional networks.
Especially, the polyhedral domain in \cite{SinghGPV19popl} can scale up the
performance of the verifier DeepPoly to handle large networks.
Recently, there have been also efforts on combining both incomplete method (such
as abstraction)
and complete method (MILP
encoding and solving), such as the works \cite{BotoevaKKLM20} and
\cite{SinghGPV19iclr}.

All above-mentioned verification methods are actually doing qualitative
verification by considering only two cases: whether the network satisfies the
property, or not.
In most recent years, researchers have been looking into quantitative
verification to check how often a property is satisfied by a given network under
a given input distribution.
For instance, the work \cite{BalutaCMS21} examines if majority
portion of the input space still satisfies the property with a high probability.

\vspace{-1em}

\vspace{-0.2em}
\section{Conclusion}
\vspace{-0.5em}

We have proposed the block summarization and bounded back-substitution to reduce
the computational steps during back-substitution process, making it more
amenable for analyzer to handle larger network with limited computational
resources, such as having only CPU setup.
We instantiated our idea on top of DeepPoly and implement a system called
BBPoly.
Experiment shows that BBPoly can achieve the verified precision comparable to
DeepPoly but save both running time and memory allocation.
Furthermore, our system is capable of analyzing large networks with up to one
million neurons while DeepPoly cannot conclude within a decent timeout.
We believe that our proposal can assist with efficient analysis and be applied
to other methods for better scalability.

\vspace{-1em}

\vspace{-0.3em}
\section{Acknowledgement}
\vspace{-0.6em}
We are grateful to Gagandeep Singh and Mark Niklas Müller for their prompt and patient answer to our queries on DeepPoly/GPUPoly.
This research is supported by a Singapore Ministry of Education Academic Research Fund Tier 1 T1-251RES2103. 
The second author is supported by both a Singapore Ministry of Education Academic Research Fund Tier 3 MOE2017-T3-1-007 and a Singapore National Research Foundation Grant R-252-000-B90-279 for the project Singapore Blockchain Innovation Programme.

\vspace{-0.7em}

\bibliographystyle{unsrt}
\bibliography{refe}


\appendix
\newpage

\section{Concrete Bound Computation from Symbolic Constraints}
\label{appendix:concre_bound_computation}

As can be seen from the illustrative example in \autoref{sec:abInt_on_egNet}, the symbolic constraints are used to compute the concrete bounds of affine neurons and this is the stratagem proposed in DeepPoly to achieve  tighter intervals. To do so, it requires us to evaluate the
minimum or maximum value for the symbolic constraints.

For symbolic lower constraint of neuron $m$, suppose we have $m \geq w_1\cdot
x_1 + \dots + w_n\cdot x_n$.
Note that $x_1, \dots, x_n$ represent the neurons from some preceding layer.
Since the analysis of the preceding neurons has completed, we know the
concrete bounds for each $x_i \in \{x_1, \dots, x_n\}$ and have $x_i \in [l_i,
  u_i]$.

To evaluate the expression $w_1\cdot x_1 + \dots + w_n\cdot x_n$ for lower
bound computation, we will calculate the minimum value $\beta$ of $w_1\cdot x_1
+ \dots + w_n\cdot x_n$.
The minimum value is computed by the following rules:

\begin{itemize}
\item For \textit{positive} coefficient $w_i$, replace $x_i$ with the
  concrete \textit{lower} bound $l_i$.

\item For \textit{negative} coefficient $w_i$, replace $x_i$ with the
  concrete \textit{upper} bound $u_i$.
\end{itemize}

Therefore, we have $m \geq w_1\cdot x_1 + \dots + w_n\cdot x_n \geq \beta$,
and $\beta$ functions as the concrete lower bound of neuron $m$.
%
Similarly for the upper bound computation of $m$, we have symbolic constraint $m \leq
w_1\cdot x_1 + \dots + w_n\cdot x_n$ and calculate the maximum value $\delta$ of
$w_1\cdot x_1 + \dots + w_n\cdot x_n$ by:

\begin{itemize}

\item For \textit{positive} coefficient $w_i$, replace $x_i$ with the
  concrete \textit{upper} bound $u_i$

\item For \textit{negative} coefficient $w_i$, replace $x_i$ with the
  concrete \textit{lower} bound $l_i$

\end{itemize}

So we have $m \leq w_1\cdot x_1 + \dots + w_n\cdot x_n \leq \delta$ and $\delta$
is the concrete upper bound of neuron $m$.

\section{Explanation why a block is ended by an affine layer}
\label{appendix:explain-layer-end-affine}
Without loss of generality, suppose the symbolic lower constraint for a neuron $m$ at a layer is expressed as a linear combination of neurons $\vec x_i$ at a preceding layer: $m \geq w_1\cdot x_1 + \dots + w_n\cdot x_n$. In a back-substitute step, $\vec x_i$ will be replaced by its own symbolic lower constraint expressed in terms of its preceding layer: $x_i \geq u_{i1} \cdot y_1 + \dots + u_{in}\cdot y_n$. Expressing the symbolic lower constraint of $m$ in terms of $\vec y_j$ would require $n^2$ computation for combining  coefficients $\vec w_i$ and $\vec u_{ij}$. On the other hand, some of these combining computations can be eliminated if the ReLU neurons connecting neuron $m$ and layer $\vec x_i$ can be set to 0. As it is not infrequent for ReLU neuron values to be set to 0 (for its symbolic lower constraint), by letting a block to end at an affine layer and begin at an ReLU layer, we increase the opportunity to leverage on the symbolic lower constraint of the ReLU neurons being 0, which will lead to elimination of $n^2$ coefficient computation when back propagating from one block to another.
Thus, to preserve the back-substitution approach in DeepPoly and speed up our back-substitution over block summary, we elect to end summary block with an affine layer.

\section{Empirical proof of bounded back-substitution heuristic}
\label{appendix:early_termi_appendix}
To empirically test the idea of bounded back-substitution, we experiment on the networks enumerated in \autoref{tab:net_for_early_term}, where the classified dataset, the layer number and the number of hidden neurons are listed as well. Please be noted that one actual layer in the network corresponds to one affine layer and one ReLU layer in the analysis. Therefore networks with 9 layers totally have 18 layers in DeepPoly's representation, so setting \textit{max\_backsub\_steps} as 18 meaning that we will NOT bound the back-substitution process and it will be equal to DeepPoly method.

\begin{table}[ht]
\caption{Networks for early termination experiment}
\centering
\begin{tabular}{|l|l|l|l|l|}
\hline
Network \ & Type \ &Dataset \ &  \#Layer \ &  \#hidden neurons \ \\
\hline
mnist\_9\_200 & fully connected \ &  MNIST & 9 & 1,610\\
\hline
mnist\_9\_100 & fully connected & MNIST & 9 & 810\\
\hline
mnist\_ffcnRELU\_Point\_6\_500 \  & fully connected & MNIST & 6 & 3,000\\
\hline
mnist\_6\_100 & fully connected & MNIST & 6 & 510\\
\hline
convSuperRELU\_mnist & convolutional& MNIST & 6 & 88,544\\
\hline
convBigRELU\_mnist & convolutional& MNIST & 6 & 48,064\\
\hline
cifar\_6\_100 & fully connected & CIFAR10 & 6 & 610\\
\hline
convBigRELU\_cifar & convolutional & CIFAR10 \ & 6 & 62,464\\
\hline
\end{tabular}
\label{tab:net_for_early_term}
\end{table}

\begin{figure*}[!ht]
           \begin{subfigure}[b]{0.495\textwidth}
            \centering
            \includegraphics[width=\textwidth]{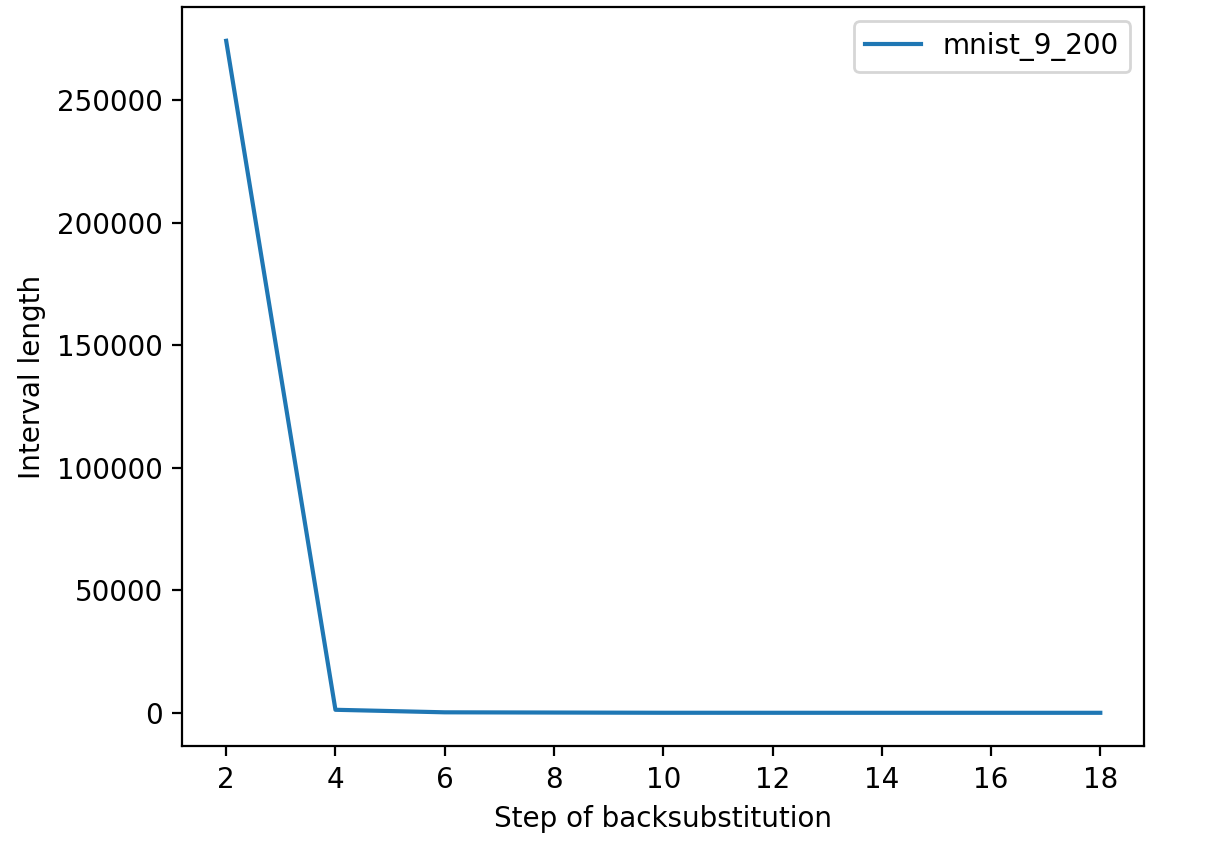}
            \caption{}%
        \end{subfigure}
        \hfill
        \begin{subfigure}[b]{0.495\textwidth}
            \centering
            \includegraphics[width=\textwidth]{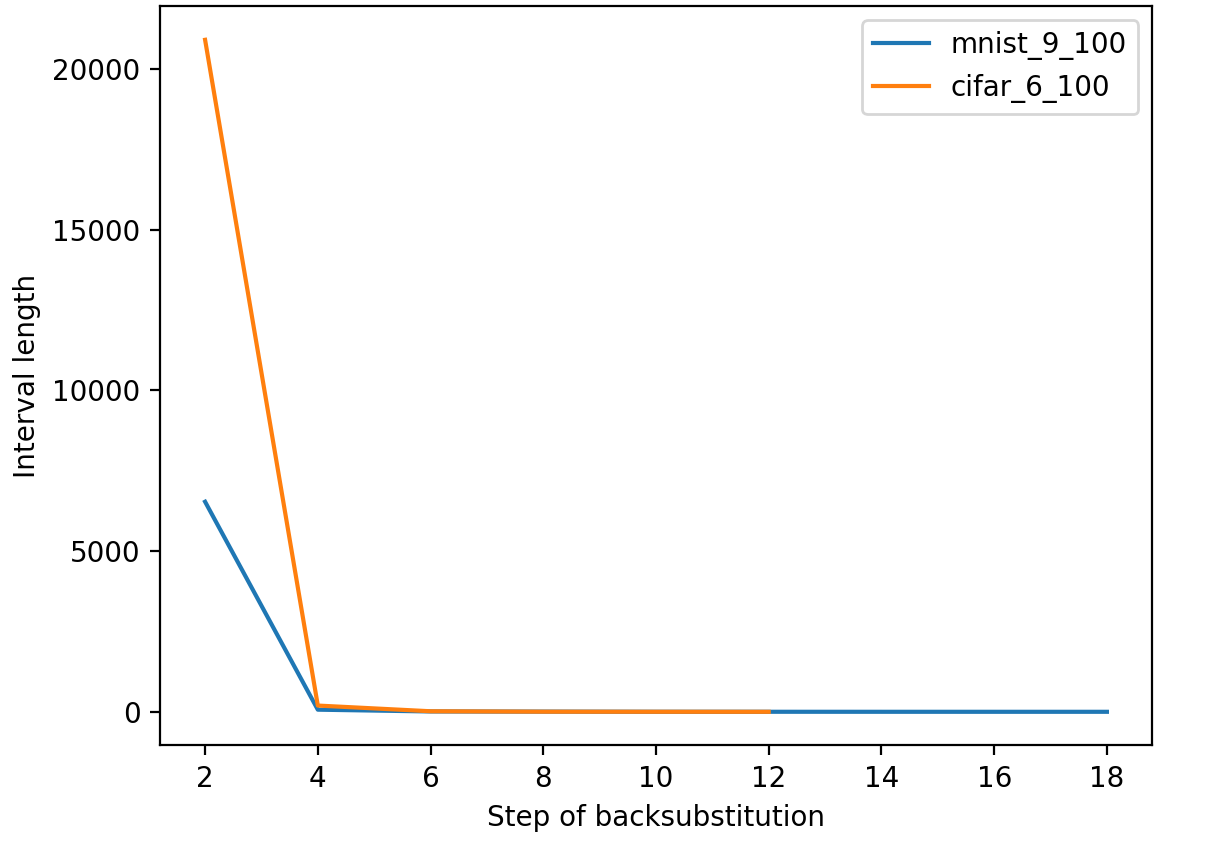}
            \caption{}%
        \end{subfigure}
        \vskip\baselineskip
        \begin{subfigure}[b]{0.495\textwidth}
            \centering
            \includegraphics[width=\textwidth]{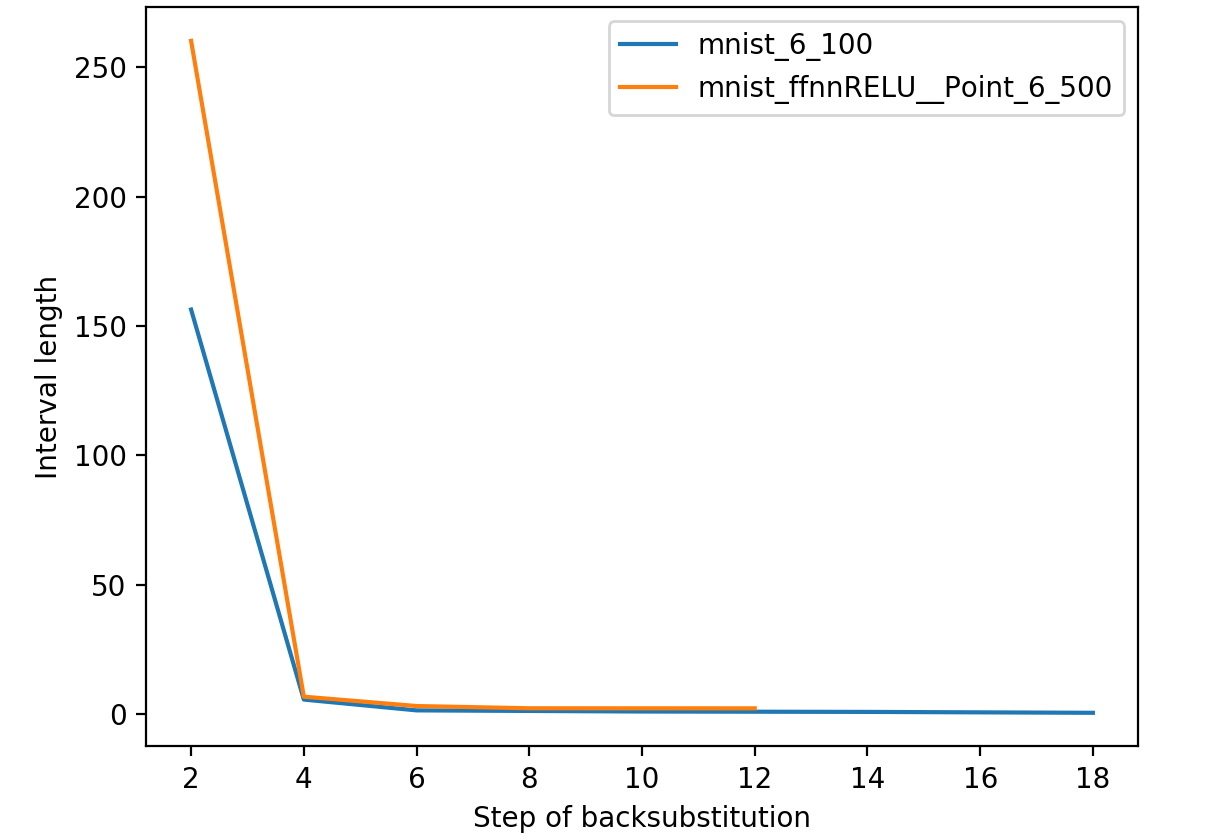}
            \caption{}%
        \end{subfigure}
        \hfill
        \begin{subfigure}[b]{0.495\textwidth}
            \centering
            \includegraphics[width=\textwidth]{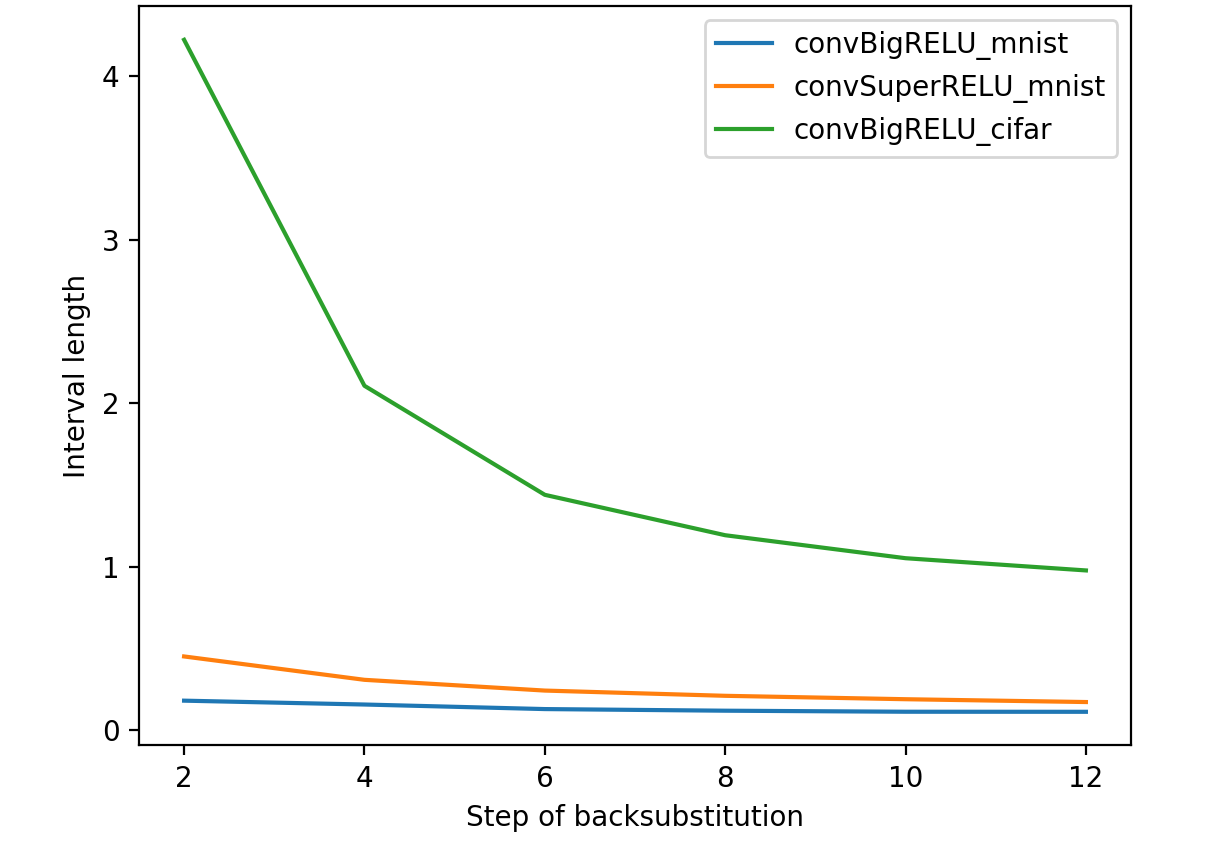}
            \caption{}%
        \end{subfigure}
\caption{Empirical result of bounded back-substitution}
\label{fig:early_term_expris}
\end{figure*}

\autoref{fig:early_term_expris} demonstrates the tendency of interval length of output neuron with respect to different setting of \textit{max\_backsub\_steps}. The interval length is averaged over different input images, different perturbations applied and different output neurons. From the outline, we can see that as we allow deeper back-substitution, the benefit will become less and less significant.

\section{Perturbation size for fully-connected and convolutional networks}
\label{appendix:perturbation-size}
We set six different values for perturbation size $\epsilon$ according to the following settings made in DeepPoly paper:
\begin{itemize}
    \item $\epsilon \in \{0.005, \ 0.01, \ 0.015, \ 0.02, \ 0.025, \ 0.03\}$ for MNIST fully-connected networks;
    \item $\epsilon \in \{0.02, \ 0.04, \ 0.06, \ 0.08, \ 0.1, \ 0.12
\}$ for MNIST convolutional networks;
    \item $\epsilon \in \{0.0002, \ 0.0004, \ 0.0006, \ 0.0008, \ 0.001, \ 0.0012
\}$ for CIFAR10 fully-connected networks;
\item $\epsilon \in \{0.002, \ 0.004, \ 0.006, \ 0.008, \ 0.01, \ 0.012
\}$ for CIFAR10 convolutional networks.
\end{itemize}
Note that, since the specified $\epsilon$ set couldn't really differentiate between various $\epsilon$ and methods for MNIST convolutional big network as demonstrated in \autoref{fig:mnist_convbig_prec}, so we choose a larger set $\epsilon \in \{0.12, \ 0.14, \ 0.16, \ 0.18, \ 0.2, \ 0.22
\}$ for MNIST convolutional super network.

\section{Precision and time comparison in table}
\label{appendix:precision_in_tables}
We also demonstrate the precision and average execution time per image in tables. The experimental result of fully-connected and convolutional networks for MNIST/CIFAR10 is given in \autoref{tab:precision_conv_ffcn}. In the table, we record the number of verified images out of 100 test images together with the average execution time per image (which is recorded within the square brackets) for different experiment settings and three different methods. The three methods include

\begin{itemize}
\item DeepPoly;
\item Our method with block summarization, which is denoted as ``Block\_sum";
\item Our summary-over-input method, denoted as ``Input\_sum".
\end{itemize}

\begin{table}[!ht]
  \caption{Precision for convolutional and fully-connected networks}
  \vspace{1em}
  \centering
  \begin{tabular}{|c|c|c|c|c|}
    \hline
    \multirow{3}{*}{\makecell{\textbf{Neural} \\[-0.2em] \textbf{Network}}} &
    \multirow{3}{*}{\makecell{\textbf{Perturbation} \\[-0.2em] \textbf{Size}}} &
    \multicolumn{3}{c|}{\textbf{\#Verified Images out of 100 Images and Time(s)}} \\

    \cline{3-5}
    & &
    \hspace{1.5em}\textbf{DeepPoly}\hspace{1.5em} &
    \hspace{1.5em}\makecell{\textbf{BBPoly}\\[-0.2em] (Block\_sum)}\hspace{1.5em} &
    \makecell{\textbf{BBPoly}\\[-0.2em] (Input\_sum)} \\

    \hline
    \multirow{6}{5.5em}{MNIST\\\_9\_200} & 0.005 & 92[3.299] & 74[2.773] & 66[2.644] \\
    \cline{2-5}
    & 0.01 & 67[3.598] & 18[3.170] & 12[3.068] \\
    \cline{2-5}
    & 0.015 & 29[4.011] & 3[3.544] & 2[3.368] \\
    \cline{2-5}
    & 0.02 & 8[4.466] & 1[3.838] & 0[3.566] \\
    \cline{2-5}
    & 0.025 & 3[4.720] & 0[4.026] & 0[3.751] \\
    \cline{2-5}
    & 0.03 & 2[4.897] & 0[4.135] & 0[3.851] \\
    \hline
    \multirow{6}{5.5em}{MNIST\_ffcn\\\_Point\_6\_500} & 0.005 & 99[7.244] & 99[5.494] & 99[5.336] \\
    \cline{2-5}
    & 0.01 & 98[7.357] & 97[5.452] & 97[5.467] \\
    \cline{2-5}
    & 0.015 & 95[7.673] & 94[5.634] & 94[5.622] \\
    \cline{2-5}
    & 0.02 & 89[8.039] & 82[6.065] & 82[5.966] \\
    \cline{2-5}
    & 0.025 & 71[8.754] & 58[6.668] & 55[6.717] \\
    \cline{2-5}
    & 0.03 & 49[9.545] & 38[7.489] & 37[7.390] \\
    \hline
    \multirow{6}{5.5em}{MNIST\\\_convBig\\\_RELU} & 0.02 & 95[33.96] & 95[18.87] & 95[12.40] \\
    \cline{2-5}
    & 0.04 & 95[33.12] & 95[18.74] & 95[17.76] \\
    \cline{2-5}
    & 0.06 & 95[33.68] & 95[18.85] & 95[18.31] \\
    \cline{2-5}
    & 0.08 & 95[34.21] & 95[18.92] & 95[17.37] \\
    \cline{2-5}
    & 0.1 & 94[33.55] & 93[18.59] & 93[17.96] \\
    \cline{2-5}
    & 0.12 & 93[34.06] & 92[19.38] & 91[18.21] \\
    \hline
    \multirow{6}{5.5em}{MNIST\\\_convSuper\\\_RELU} & 0.12 & 96[133.5] & 96[75.22] & 95[73.94] \\
    \cline{2-5}
    & 0.14 & 81[138.7] & 77[78.12] & 74[78.12] \\
    \cline{2-5}
    & 0.16 & 28[148.7] & 10[84.84] & 6[82.90] \\
    \cline{2-5}
    & 0.18 & 0[159.5] & 0[93.19] & 0[90.12] \\
    \cline{2-5}
    & 0.2 & 0[179.8] & 0[102.6] & 0[98.21] \\
    \cline{2-5}
    & 0.22 & 0[197.9] & 0[114.4] & 0[109.8] \\
    \hline
    \multirow{6}{5.5em}{CIFAR10\\\_ffcn\_Point\\\_6\_500} & 0.0002 & 55[22.32] & 55[14.43] & 55[15.18] \\
    \cline{2-5}
    & 0.0004 & 53[22.63] & 53[14.53] & 53[15.18] \\
    \cline{2-5}
    & 0.0006 & 53[22.75] & 52[14.76] & 49[15.08] \\
    \cline{2-5}
    & 0.0008 & 52[22.71] & 46[14.74]] & 46[15.16] \\
  \cline{2-5}
  & 0.001 & 48[22.76] & 43[14.88] & 43[15.36] \\
  \cline{2-5}
  & 0.0012 & 46[22.70] & 43[15.07] & 43[15.35] \\
  \hline
  \multirow{6}{5.5em}{CIFAR10\\\_convBig\\\_RELU} & 0.002 & 51[89.07] & 51[42.45] & 51[37.46] \\
  \cline{2-5}
  & 0.004 & 46[89.69] & 42[41.38] & 42[37.49] \\
  \cline{2-5}
  & 0.006 & 31[91.04] & 31[41.88] & 31[37.78] \\
  \cline{2-5}
  & 0.008 & 24[91.26] & 21[41.87] & 21[37.94] \\
  \cline{2-5}
  & 0.01 & 15[91.58] & 12[41.87] & 12[38.32] \\
  \cline{2-5}
  & 0.012 & 11[92.56] & 9[41.79] & 9[38.87] \\
  \hline
\end{tabular}
\label{tab:precision_conv_ffcn}
\end{table}

\section{Additional experimental result on the neural network MNIST 9x200 with different block sizes of 3, 4, 5}
\label{appendix:diff-blk}
In \autoref{tab:diff_blk}, the record ``83[2.658]" indicates that BBPoly verifies robustness for 83 images, with average execution time 2.658 seconds for each image.
\begin{table}[!h]
  \caption{Precision and execution time for different block sizes}
  \centering
  \begin{tabular}{|c|c|c|c|c|}
    \hline
    \multirow{3}{*}{\makecell{\textbf{Neural} \\[-0.2em] \textbf{Network}}} &
    \multirow{3}{*}{\makecell{\textbf{Perturbation} \\[-0.2em] \textbf{Size}}} &
    \multicolumn{3}{c|}{\textbf{\#Verified Images out of 100 Images and Time(s)}} \\

    \cline{3-5}
    & &
    \hspace{1.5em}\textbf{DeepPoly}\hspace{1.5em} &
    \hspace{1.5em}\makecell{\textbf{BBPoly}\\[-0.2em] (Block\_sum)}\hspace{1.5em} &
    \makecell{\textbf{BBPoly}\\[-0.2em] (Input\_sum)} \\
    \hline
    \multirow{6}{5.5em}{MNIST\\\_9\_200\\blk\_size=3} & 0.005 & 92[3.299] & 74[2.773] & 66[2.644] \\
    \cline{2-5}
    & 0.01 & 67[3.598] & 18[3.170] & 12[3.068] \\
    \cline{2-5}
    & 0.015 & 29[4.011] & 3[3.544] & 2[3.368] \\
    \cline{2-5}
    & 0.02 & 8[4.466] & 1[3.838] & 0[3.566] \\
    \cline{2-5}
    & 0.025 & 3[4.720] & 0[4.026] & 0[3.751] \\
    \cline{2-5}
    & 0.03 & 2[4.897] & 0[4.135] & 0[3.851] \\
    \hline
    \multirow{6}{5.5em}{MNIST\\\_9\_200\\blk\_size=4} & 0.005 & 92[3.299] & 83[2.658] & 82[2.683] \\
    \cline{2-5}
  & 0.01 & 67[3.598] & 29[3.174] & 29[3.118] \\
    \cline{2-5}
    & 0.015 & 29[4.011] & 6[3.561] & 6[3.530] \\
    \cline{2-5}
    & 0.02 & 8[4.466] & 2[3.725] & 2[3.767]\\
    \cline{2-5}
    & 0.025 & 3[4.720] & 0[3.910] & 0[3.900] \\
    \cline{2-5}
    & 0.03 & 2[4.897] &0[3.960] & 0[3.968]\\
    \hline
    
    \multirow{6}{5.5em}{MNIST\\\_9\_200\\blk\_size=5} & 0.005 & 92[3.299] & 88[2.617] & 88[2.643] \\
    \cline{2-5}
    & 0.01 & 67[3.598] & 37[3.041] & 37[3.047] \\
    \cline{2-5}
    & 0.015 & 29[4.011] & 9[3.504] & 9[3.473] \\
    \cline{2-5}
    & 0.02 & 8[4.466] &2[3.712] & 2[3.729] \\
    \cline{2-5}
    & 0.025 & 3[4.720] & 0[3.831] & 0[3.854] \\
    \cline{2-5}
    & 0.03 & 2[4.897] & 0[3.930] & 0[3.940] \\
    \hline
\end{tabular}
\label{tab:diff_blk}
\end{table}

\section{Algorithm of analysis with summary-over-input method}
\label{appendix:algo-sum-input}
We present the overall procedure of analysis with summary-over-input in \autoref{algo:suminputalgo}. The algorithm is very similar to \autoref{algo:methodalgo}, the differences between the two algorithms are:
\begin{itemize}
    \item The bounded back-substitution heuristic is not involved in summary-over-input method;
    \item In the summary-over-input method, the summary is directly defined over the actual input layer of the network in stead of the start layer of the given block. Therefore, after back-substitution with summary, the new constraint will directly be defined over input layer (line 13-14);
    \item If $\gamma_k$ is the end layer of a block and the current preceding layer $\gamma_{pre}$ is the input layer of the network, we then store the current $\Upsilon_k$ as the block summary of layer $\gamma_{k}$ (line 19-20).
\end{itemize}

\begin{algorithm}
\textbf{Input:}
  $M$ is the network (eg. \autoref{fig:HypoNet});\\
  $\segmpara$ is the network segmentation parameter

  \textbf{Annotatation:}
   input layer of $M$ as $\gamma_{in}$;\\ constraint set of affine layer $\gamma_{k}$ as $\Upsilon_k$; \\
   the set of concrete bounds for neurons in layer $\gamma_{k} \in
  M$ as $C_k$; \\
  the \textit{segmented} network model as  $\mathcal{M}$\\

  \textbf{Assumption:}
  the analysis is conducted in ascending order of the layer index\\
  \textbf{Output:}
  tightest concrete bounds $C_k$ computed for all layer $\gamma_{k} \in M$

\begin{algorithmic}[1]
  \STATE $\mathcal{M} \gets \Call{SegmentNetwork}{M,\segmpara}$
  \FORALL{layer $\gamma_k \in \mathcal{M}$ }
  \IF{$\Call{IsReluLayer}{\gamma_k}$}
    \STATE  $\gamma_{pre} \gets \Call{PredecessorLayer}{\gamma_k}$
    \STATE $C_k \gets \Call{ComputeReluLayer}{\gamma_{pre}}$
    \ELSE
    \STATE $\gamma_{pre} \gets \Call{PredecessorLayer}{\gamma_k}$

    \STATE $\Upsilon_k \gets \Call{GetSymbolicConstraints}{\gamma_k}$
    \STATE $C_k \gets \Call{EvaluateConcreteBounds}{\Upsilon_k, \gamma_{pre}}$
    \WHILE{$\gamma_{pre} \neq  \gamma_{in}$}
      \IF{$\Call{IsEndLayer}{\gamma_{pre}}$ \algorithmiccomment{$\gamma_{pre}$ is the end layer of a block}
        }
        \STATE $sum \gets \Call{ReadSummary}{\gamma_{pre}}$
        \STATE $\Upsilon_k \gets \Call{BacksubWithBlockSummary}{\Upsilon_k, sum}$
        \STATE $\gamma_{pre} \gets \gamma_{in}$
      \ELSE
        \STATE $sym\_cons \gets \Call{GetSymbolicConstraints}{\gamma_{pre}}$
        \STATE $\Upsilon_k \gets \Call{BacksubWithSymbolicConstraints}{\Upsilon_k, sym\_cons}$
        \STATE $\gamma_{pre} \gets \Call{PredecessorLayer}{\gamma_{pre}}$
    \ENDIF
    \IF{$\Call{IsEndLayer}{\gamma_k}$
            \AND $\gamma_{pre} = \gamma_{in} $}
          \STATE $\Call{StoreSummary}{\gamma_k, \Upsilon_k}$
    \ENDIF
    \STATE $temp\_ck \gets \Call{EvaluateConcreteBounds}{\Upsilon_k, \gamma_{pre}}$
    \STATE $C_k \gets \Call{UpdateBounds}{C_k, temp\_ck}$
  \ENDWHILE
  \ENDIF
\ENDFOR
\RETURN all $C_k$ for all layer $\gamma_k \in \mathcal{M}$
\end{algorithmic}
\caption{Overall analysis procedure with summary-over-input}
\label{algo:suminputalgo}
\end{algorithm}


\end{document}